\documentclass[aps, pre, a4paper, floatfix,  twocolumn, longbibliography]{revtex4-1}

\usepackage[utf8]{inputenc}
\usepackage{epsfig}
\usepackage[T1]{fontenc}
\usepackage[english]{babel}
\usepackage[table]{xcolor}
\usepackage{t1enc}
\usepackage{graphicx}
\usepackage{amssymb}
\usepackage{amsmath}
\usepackage{relsize}
\usepackage[percent]{overpic}
\usepackage{bm}
\usepackage{hyperref}
\usepackage[section]{placeins}

\usepackage[normalem]{ulem}

\newcommand{\ceil}[1]{{\lceil #1\rceil}}
\newcommand{\dd}{\mathrm{d}}
\newcommand{\ee}{\mathrm{e}}

\usepackage{mathtools}
\def\multiset#1#2{\ensuremath{\left(\kern-.3em\left(\genfrac{}{}{0pt}{}{#1}{#2}\right)\kern-.3em\right)}}

\usepackage{amsmath}

\newcommand{\bb}{\bm{b}}
\newcommand{\A}{\bm{A}}
\newcommand{\x}{\bm{x}}

\usepackage{verbatim}
\usepackage{overpic}
\usepackage{booktabs}
\usepackage{placeins}

\begin{document}

\title{Nonparametric weighted stochastic block models}

\author{Tiago P. Peixoto}
\email{t.peixoto@bath.ac.uk}
\affiliation{Department of Mathematical Sciences and Centre for Networks
and Collective Behaviour, University of Bath, Claverton Down, Bath BA2
7AY, United Kingdom}
\affiliation{ISI Foundation, Via Alassio 11/c, 10126 Torino, Italy}

\begin{abstract}
  We present a Bayesian formulation of weighted stochastic block models
  that can be used to infer the large-scale modular structure of
  weighted networks, including their hierarchical organization. Our
  method is nonparametric, and thus does not require the prior knowledge
  of the number of groups or other dimensions of the model, which are
  instead inferred from data. We give a comprehensive treatment of
  different kinds of edge weights (i.e. continuous or discrete, signed
  or unsigned, bounded or unbounded), as well as arbitrary weight
  transformations, and describe an unsupervised model selection approach
  to choose the best network description. We illustrate the application
  of our method to a variety of empirical weighted networks, such as
  global migrations, voting patterns in congress, and neural connections
  in the human brain.
\end{abstract}

\maketitle

\section{Introduction}

Many network systems lack a natural low-dimensional embedding from which
we can readily extract their most prominent large-scale
features. Instead, we have to infer this information from data,
typically by decomposing the observed network into
modules~\cite{fortunato_community_2010}. A principled approach to
perform this task is to formulate generative models that allow this
modular decomposition to be found via statistical
inference~\cite{peixoto_bayesian_2017}. The most fundamental model used
for this purpose is the stochastic block model
(SBM)~\cite{holland_stochastic_1983}, which groups nodes according to
their probabilities of connection to the rest of the network. However, a
central limitation of most SBM implementations is that they are defined
strictly for simple or multigraphs. This means that they do not
incorporate extra information on the edges, which are typically present
in a variety of systems, and are required for an accurate representation
of their structure. For example, to the existence of a route between two
airports is associated a distance, to the biomass flow between two
species in a food web is associated a flow magnitude, etc. In this work,
we develop variations of the SBM that allow for this type of
information on the edges to be incorporated into the network model and
guide the partition of the nodes into groups in a statistically
meaningful way.

We follow the same basic idea put forth by Aicher et
al.~\cite{aicher_learning_2015}, who adapted the SBM to weighted
networks by including edge values as additional covariates. However, our
approach diverges from Ref.~\cite{aicher_learning_2015} in key
aspects. First, here we develop a nonparametric Bayesian approach, based
on exact integrated likelihoods, that is capable of inferring the
dimension of the model ---
e.g. the number of groups --- from the data itself, without requiring it
to be known \emph{a priori}. This is achieved by departing from the
canonical exponential family of distributions, and using instead
\emph{microcanonical} formulations that are easier to compute exactly
and approach the canonical distributions asymptotically. Second, our
approach also infers the hierarchical modular structure of the network,
extending the nested SBM of
Refs.~\cite{peixoto_hierarchical_2014,peixoto_nonparametric_2017} to the
weighted case. The hierarchical nature of the model is implemented via
structured Bayesian priors that have been shown to significantly
decrease the tendency of the nonparametric approach to
underfit~\cite{peixoto_parsimonious_2013} and are capable of uncovering
small but statistically significant modules in large
networks~\cite{peixoto_hierarchical_2014,peixoto_nonparametric_2017}.
And third, our approach is efficient, making use of MCMC sampling that
requires only $O(E)$ operations per sweep, where $E$ is the number of
edges in the network, independently of the number of groups.

This paper is organized as follows. In Sec.~\ref{sec:wsbm} we present
our general approach, and in Sec.~\ref{sec:empirical} we illustrate its
use in a variety of empirical weighted network datasets. In
Sec.~\ref{sec:likelihoods} we elaborate on the diverse models for edge
weights based on basic properties (such as whether they are discrete or
continuous, signed or unsigned, bounded or unbounded), show how these
models can be extended via weight transformations, and how different
models can be chosen via Bayesian model selection. We finalize in
Sec.~\ref{sec:conclusion} with a discussion.

\section{Weighted SBMs via edge covariates}\label{sec:wsbm}

We consider generative models for networks that, in addition to the
adjacency matrix $\A=\{A_{ij}\}$, also possess real or discrete edge
covariates $\x=\{\x_{ij}\}$ on the edges. Without loss of generality,
here we assume that the networks are multigraphs,
i.e. $A_{ij}\in\mathbb{N}_0$, such that $\x_{ij}$ is a vector containing
one weight for each parallel edge between nodes $i$ and $j$, and no
weights if $A_{ij}=0$. Furthermore, we assume that the edge existence is
decoupled from its weight, i.e. the non-existence of an edge is
different from an edge with zero weight (the special case where the
zeros of the adjacency matrix are considered values of the edge
covariates can be recovered by using a complete graph in place of $\A$,
and adapting $\x$ accordingly). As done in
Ref.~\cite{aicher_learning_2015}, we follow the underlying assumption of
the SBM that the nodes are divided into $B$ groups, with
$b_i\in\{1,\dots,B\}$ specifying the group membership of node $i$, and
where in addition to the edge placement, the edge weights are sampled
only according to the group memberships of their endpoints. Concretely,
this means they are both sampled from parametric distributions that are
conditioned only on the group memberships of the nodes i.e.
\begin{equation}
  P(\A,\x|\bm{\theta},\bm{\gamma},\bb) =
  P(\x|\A,\bm{\gamma},\bb)P(\A|\bm{\theta},\bb)
\end{equation}
with the covariates being sampled only on existing edges,
\begin{align}
  P(\x | \A, \bm{\gamma}, \bb) &= \prod_{r\le s}P(\bm{x}_{rs}|\bm{\gamma}_{rs})
\end{align}
with $\bm{x}_{rs} = \{\bm{x}_{ij}|A_{ij}>0\land (b_i,b_j)=(r,s)\}$ being
the covariates between groups $r$ and $s$, and where $\bm\gamma_{rs}$ is
a set of parameters that govern the sampling of the weights between
groups $r$ and $s$. The placement of the edges is done independently of
the weights by choosing any SBM flavor with parameters $\bm{\theta}$;
For example, with the degree-corrected SBM~\cite{karrer_stochastic_2011}
we would have
\begin{align}
  P(\A|\bm{\theta}=\{\bm{\lambda},\bm{\kappa}\},\bb) =
  \prod_{i<j}\frac{\ee^{-\lambda_{b_i,bj}\kappa_i\kappa_j}
    (\lambda_{b_i,bj}\kappa_i\kappa_j)^{A_{ij}}}{A_{ij}!},
\end{align}
where $\lambda_{rs}$ controls the number of edges that are placed
between groups, and $\kappa_i$ the expected degree of node $i$.

Given the generative model above, we could proceeded by estimating the
parameters $\bm{\theta}$ and $\bm{\gamma}$ via maximum
likelihood. However, doing so would be subject to overfitting, as the
likelihood would increase monotonically with the complexity of the
model.  Instead, here we are interested in solving a more general and
arguably more well-posed problem, namely to obtain the Bayesian
posterior probability of partitions, in a \emph{nonparametric} manner,
taking into account only the weighted network,
\begin{equation}\label{eq:posterior_b}
  P(\bb|\A,\x) = \frac{P(\A,\x|\bb)P(\bb)}{P(\A,\x)},
\end{equation}
where the numerator contains the marginal likelihood integrated over the
model parameters
\begin{align}
  P(\A,\x|\bb) &= \int P(\A,\x|\bm{\theta},\bm{\gamma},\bb)P(\bm{\theta})P(\bm{\gamma})\,\dd\bm\theta\dd\bm\gamma \nonumber\\
                          &= P(\A|\bb)P(\x | \A,\bb),\label{eq:joint_ax}
\end{align}
and where
\begin{equation}\label{eq:marg_sbm}
  P(\A|\bb) = \int P(\A|\bm{\theta},\bb)P(\bm{\theta})\,\dd\bm\theta
\end{equation}
is the marginal likelihood of the unweighted network integrated over the
relevant parameters. Integrated marginal likelihoods of this kind were
considered in numerous works for several unweighted model
variants~\cite{guimera_missing_2009,yan_active_2010,peixoto_parsimonious_2013,
  peixoto_hierarchical_2014,come_model_2015,newman_estimating_2016,
  peixoto_nonparametric_2017}. In this work, our approach is fully
independent of any particular choice made for this part of the
model. However, in our experiments we will use the nested microcanonical
degree-corrected SBM described in Ref.~\cite{peixoto_hierarchical_2014,
peixoto_nonparametric_2017}, due to its efficient and multi-scale
nature, as well as a much reduced tendency to underfit when used with
large networks. Furthermore, its hierarchical nature will allow us to
describe summaries of the network
--- taking into accounts its edge covariates --- at multiple scales,
providing a bird's-eye view of large datasets. We use this model without
sacrificing generality, since the usual non-hierarchical SBM amounts
exactly to using the nested version with just one hierarchical level.

The crucial part in Eq.~\ref{eq:joint_ax} that completes our
nonparametric approach is the marginal likelihood of the edge weights,
which is integrated over the weight parameters $\bm{\gamma}$ according
to their prior distribution $P(\bm\gamma_{rs})$, which is the same for
every pair of groups $r$ and $s$,
\begin{align}
  P(\x | \A,\bb) &= \int P(\x | \A, \bm{\gamma}, \bb)P(\bm{\gamma})\,\dd\bm{\gamma} \nonumber\\
  &= \prod_{r\le s}\int P(\bm{x}_{rs}|\bm\gamma_{rs})P(\bm\gamma_{rs})\,\dd\bm\gamma_{rs}.\label{eq:marg_w}
\end{align}
The form of the prior distribution $P(\bm\gamma)$ is usually conditioned
on hyperparameters $\bm\eta$, which represent our \emph{a priori}
assumptions about the data. In order for our inference approach to
retain its nonparametric character, we need these hyperparameters to
take a single global value, i.e.
$P(\bm\gamma_{rs})=P(\bm\gamma_{rs}|\bm\eta)$ for all groups $r$ and
$s$. Alternatively, we may treat $\bm\eta$ as latent variables, and
sample them from their own distribution, $P(\bm\eta)$, thereby reducing
the sensitivity to our \emph{a priori} assumptions. This idea fits well
with the nested version of the SBM we will be
using~\cite{peixoto_hierarchical_2014, peixoto_nonparametric_2017},
which, as part of its prior probabilities, considers the groups
themselves as nodes of a smaller multigraph that is also generated by
the SBM, with its nodes put in their own groups, forming an even smaller
multigraph, and so on recursively, following a nested hierarchy
$\{\bb^l\}=\{\{b_r^{(l)}\}_l\}$, so that $b_r^{(l)} \in \{1,\dots,B_l\}$
is the group membership of group/node $r$ at the hierarchy level
$l\in\{1, \dots, L\}$, with the boundary condition that the number of
groups at the topmost level $l=L$ is $B_L=1$ (see Fig. 1 in
Ref.~\cite{peixoto_hierarchical_2014} for an illustration of the
generative process). Therefore, the adjacency of the multigraph at level
$l$ is
\begin{equation}\label{eq:hadjacency}
  m_{rs}^l = \sum_{tu}\frac{m_{tu}^{l-1}\delta_{b^{(l)}_t,r}\delta_{b^{(l)}_u,s}}{\delta_{rs}+1},
\end{equation}
where we assume $m_{ij}^0=A_{ij}$. Following the same logic, we may
consider the parameters $\bm\gamma$ as edge covariates in the multigraph
of groups, which themselves are generated by another model in a level
above, and so on. We may thus let $\bm\gamma^1\equiv\bm\gamma$ and
$\bm\gamma^2\equiv\bm\eta$ be the first two levels of a hierarchical
model, given recursively by
\begin{equation}
  P(\bm{\gamma}^l|\A,\bb^{l+1},\bm\gamma^{l+1}) =
  \prod_{t\le u}P(\bm{\gamma}_{tu}^{l}|\A,\bb^{l+1},\bm\gamma^{l+1}_{b_t^{(l+1)},b_u^{(l+1)}}),
\end{equation}
where $\bm{\gamma}_{tu}^{l+1} = \{\bm{\gamma}_{rs}^l|m_{rs}^l>0\land
(b_r^{(l+1)},b_s^{(l+1)})=(t,u)\}$ are the hyperparameters between
groups $(t,u)$ at level $l+1$, with $m_{rs}^l$ given by
Eq.~\ref{eq:hadjacency}. The final model is then obtained by integrating
over the entire hierarchy,
\begin{multline}\label{eq:marg_w_nested}
  P(\x | \A,\{\bb^l\}) = \\
  \int P(\x | \A, \bm{\gamma}^1, \bb^1)\prod_{l=1}^LP(\bm{\gamma}^l|\A,\bm\gamma^{l+1},\bb^{l+1})\;\dd\bm{\gamma}^l,
\end{multline}
assuming the boundary condition $\bm{\gamma}^{L+1}=\{\hat{\bm{\gamma}}\}$,
such that $\hat{\bm{\gamma}}$ is a single set of hyperparameters that
are left out of the integration at the topmost level, reflecting only
global aspects of the covariates, without a significant effect on the
model structure and dimension. Instead of defining a unique model, we
will consider a variety of elementary choices for $P(\x|\bm\gamma)$ and
$P(\bm\gamma)$ that reflect the precise nature of the covariates
(e.g. continuous or discrete, signed or unsigned, bounded or unbounded),
and for which Eq.~\ref{eq:marg_w_nested} can be computed exactly. In
particular, we will make use of microcanonical formulations of the
weight distributions that permit the straightforward computation of the
integrals, without sacrificing descriptive power.  We leave the
derivations of the likelihood expressions for
Sec.~\ref{sec:likelihoods}, and we proceed with a general outline, and
an analysis of this approach for empirical networks.

When using the nested model, we have a posterior distribution over
hierarchical partitions,
\begin{equation}\label{eq:posterior_b_l}
  P(\{\bb^l\}|\A,\x) = \frac{P(\A,\x|\{\bb^l\})P(\{\bb^l\})}{P(\A,\x)},
\end{equation}
which can be marginalized, if we so desire, to obtain only the partition
at the bottom level $\bb\equiv\bb^1$,
\begin{equation}
  P(\bb|\A,\x) = \sum_{\{\bb^l|\,l>1\}}P(\{\bb^l\}|\A,\x).
\end{equation}
However, most typically we will want to obtain the entire hierarchical
partition, as it is useful for a multilevel description of the data.
Since the posterior in Eq.~\ref{eq:posterior_b_l} involves a prior
probability of the partition $P(\{\bb^l\})$ (described in detail in
Ref.~\cite{peixoto_nonparametric_2017}), and is integrated over all
remaining model parameters, it possesses an inherent
\emph{regularization} property, where overly complicated models are
penalized with a lower posterior
probability~\cite{jaynes_probability_2003}. This means that, differently
from maximum likelihood approaches, we can infer properties related to
the size of the model, such as the number of groups $B$ and hierarchy
depth $L$, without danger of overfitting. Furthermore, as we detail
further in Sec.~\ref{sec:model-selection}, the posterior distribution
gives us a principled means of model selection according to statistical
significance, which allows us to choose the most appropriate weight
model.

Given a choice for the parametric model for weights, we compute
Eq.~\ref{eq:marg_w_nested}, which allows us to determine the posterior
distribution of the partitions in Eq.~\ref{eq:posterior_b_l} up to the
normalizing constant $P(\A,\x)$ in the denominator, which is generally
intractable. But since we cannot sample from the posterior distribution
directly even if we \emph{could} somehow compute this constant, we must
resort to MCMC importance sampling methods, for which this normalizing
constant is luckily not needed. Since the values that need to be
inferred are only the hierarchical labels $\{\bb^l\}$, we can use the
exact same algorithm developed for the unweighted case in
Refs.~\cite{peixoto_efficient_2014,peixoto_nonparametric_2017}, which we
summarize here. This is generally implemented by making move proposals
$\{\bb^l\}\to\{\bb^l\}'$ with probability $P(\{\bb^l\}'|\{\bb^l\})$, and
rejecting the proposal with probability $1-a$, where $a$ is the
Metropolis-Hastings~\cite{metropolis_equation_1953, hastings_monte_1970}
criterion
\begin{align}\label{eq:metropolis}
  a &= \operatorname{min}\left(1,
  \frac{P(\{\bb^l\}'|\A,\x)}{P(\{\bb^l\}|\A,\x)}\frac{P(\{\bb^l\}|\{\bb^l\}')}{P(\{\bb^l\}'|\{\bb^l\})}\right).
\end{align}
Since the ratio in Eq.~\ref{eq:metropolis} does not depend on the
normalization constant $P(\A,\x)$, the value of $a$ can be
computed exactly, and --- as long as the move proposals are ergodic ---
the algorithm above will eventually sample partitions from the desired
posterior distribution asymptotically. We can also obtain the \emph{most
likely} hierarchical partition,
\begin{align}
  \{\bb^l\}^* = \underset{\{\bb^l\}}{\operatorname{argmax}}\, P(\{\bb^l\}|\A,\x)
\end{align}
by replacing $P(\{\bb^l\}|\A,\x)\to P(\{\bb^l\}|\A,\x)^\beta$ in
Eq.~\ref{eq:metropolis} and making $\beta\to\infty$ in slow
increments. Therefore, we can both maximize and sample from the
posterior distribution, using the same algorithm. In this work we use
the same move proposals defined in Refs.~\cite{peixoto_efficient_2014,
peixoto_nonparametric_2017} where we select the layer $l$ and a node $u$
in that layer, both randomly, and use the local information of the
node's neighbourhood combined with global information on that layer to
propose a plausible move candidate for its group membership, $b^l_u\to
r$, thereby improving equilibration speed (see
Ref.~\cite{peixoto_nonparametric_2017} for details). Additionally, in order to
avoid getting trapped in metastable states, we employ the agglomerative
initialization heuristic described in Ref.~\cite{peixoto_efficient_2014}
and extended to the nested model in
Ref.~\cite{peixoto_hierarchical_2014}. The combination of these move
proposals with the likelihood of the microcanonical SBM of
Ref.~\cite{peixoto_nonparametric_2017}, as well as any of the weight
likelihoods defined in Sec.~\ref{sec:likelihoods}, yields an algorithm
where each MCMC sweep (i.e. for every node one move is attempted) is
performed in time $O(E)$, independently of how many groups are occupied
with nodes. For more details of the algorithm we defer to
Refs.~\cite{peixoto_efficient_2014, peixoto_nonparametric_2017} and to
the freely available C++ implementation in the \texttt{graph-tool}
Python library~\cite{peixoto_graph-tool_2014}.

\section{Empirical networks}\label{sec:empirical}

\subsection{Migrations between countries}

\begin{figure*}
  \begin{tabular}{ccc}
    \begin{overpic}[width=.45\textwidth]{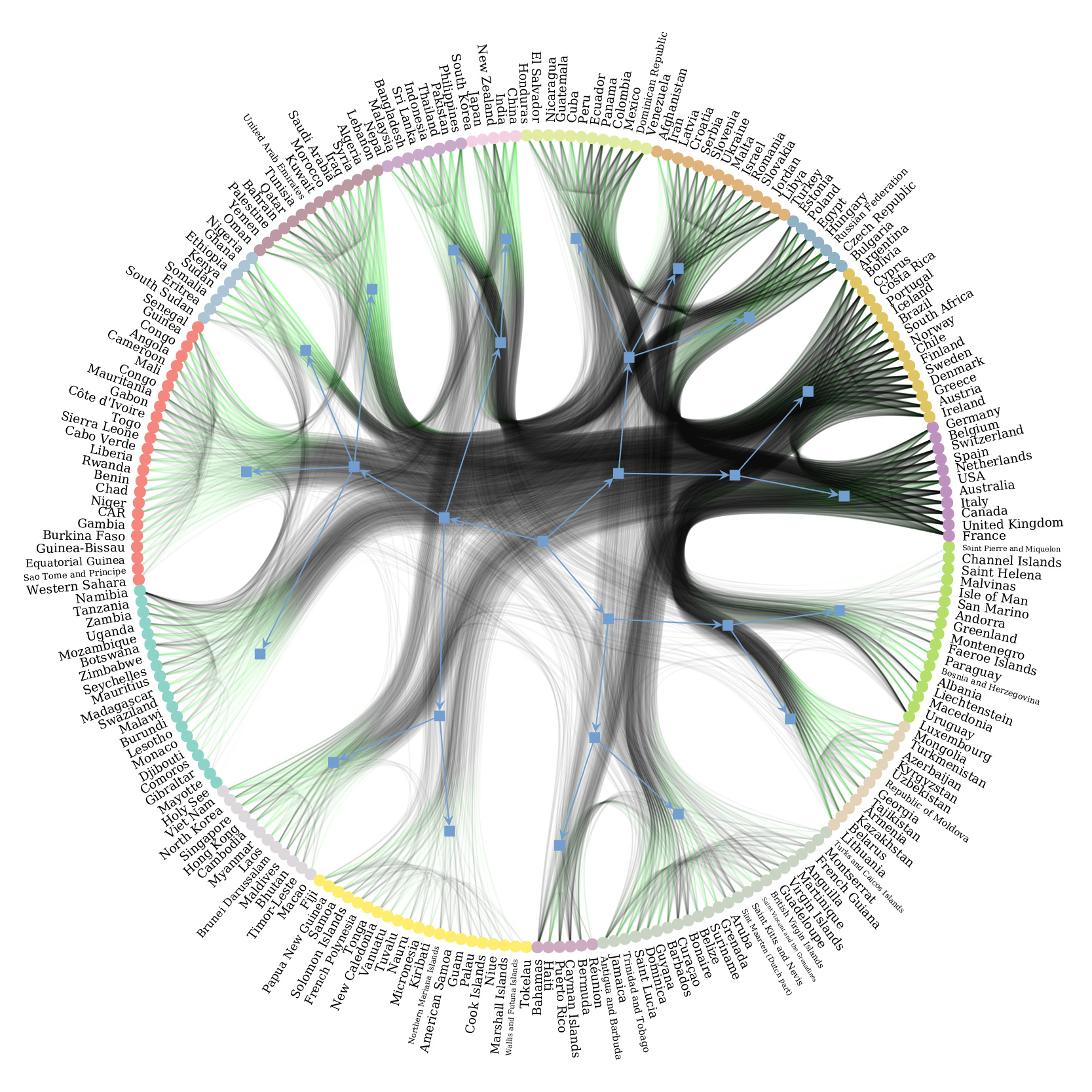}\put(0,0){(a)}\end{overpic}& 
    \begin{overpic}[width=.45\textwidth]{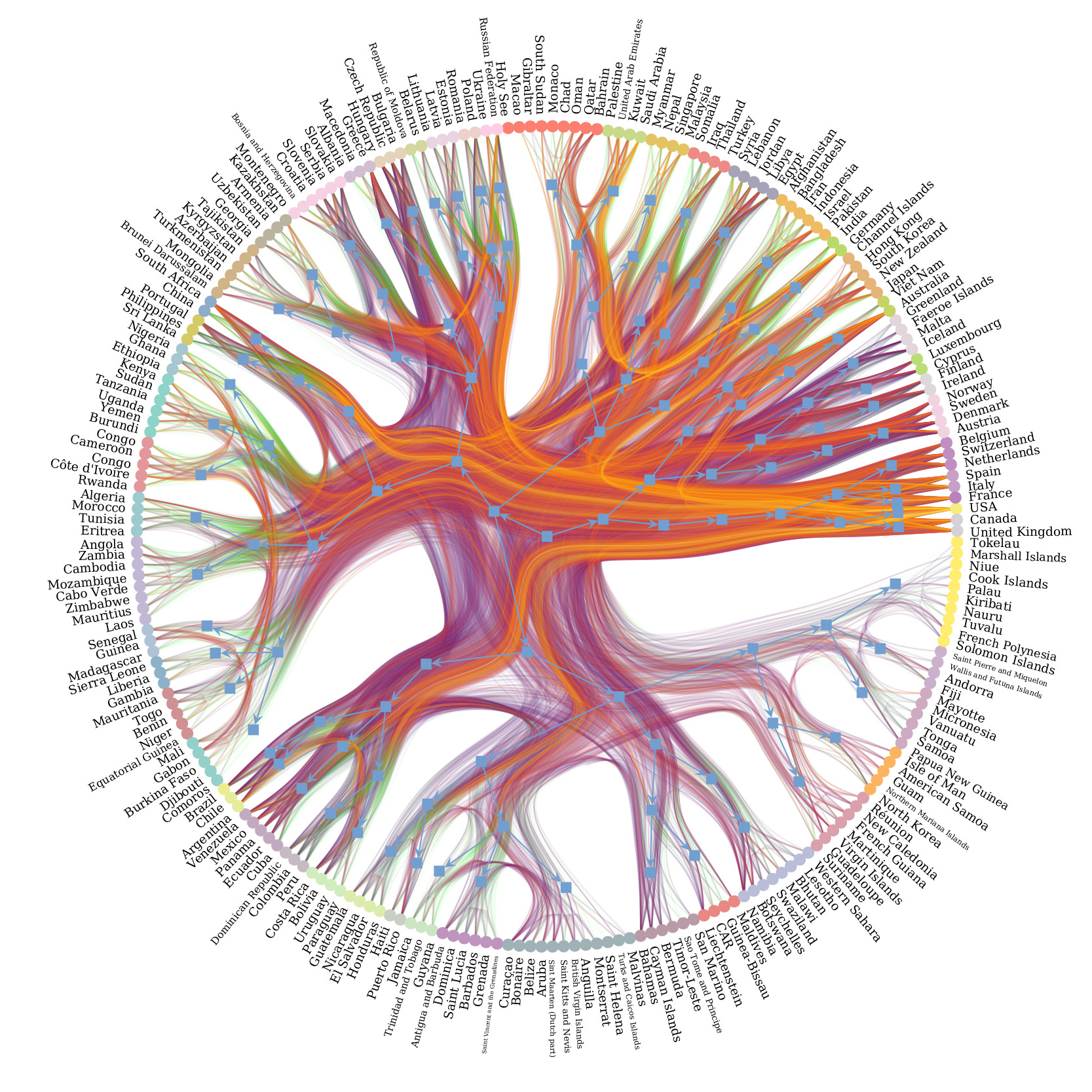}\put(0,0){(b)}\end{overpic} &
    \includegraphics[width=.4\textwidth,angle=90,origin=l,trim=0 .5cm 0 0,clip]{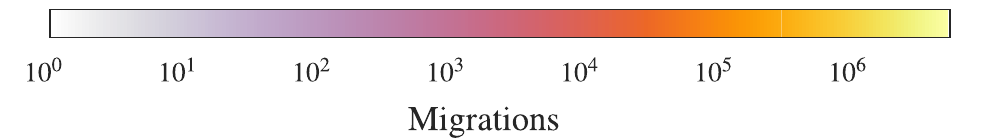}  \end{tabular}
  \caption{(a) Fit of the unweighted SBM for UN migration data, using
  the threshold approach described in the text. The edges are routed
  according the inferred hierarchy (shown in blue), using an
  edge-bundling algorithm by Holten~\cite{holten_hierarchical_2006}, and
  the edge sources are marked with a green color. (b) Fit of the
  weighted SBM for the same data with the migrant stocks included, as
  shown by the edge colors and in the legend.\label{fig:migrations}}
\end{figure*}
\begin{figure}
  \includegraphics[width=\columnwidth]{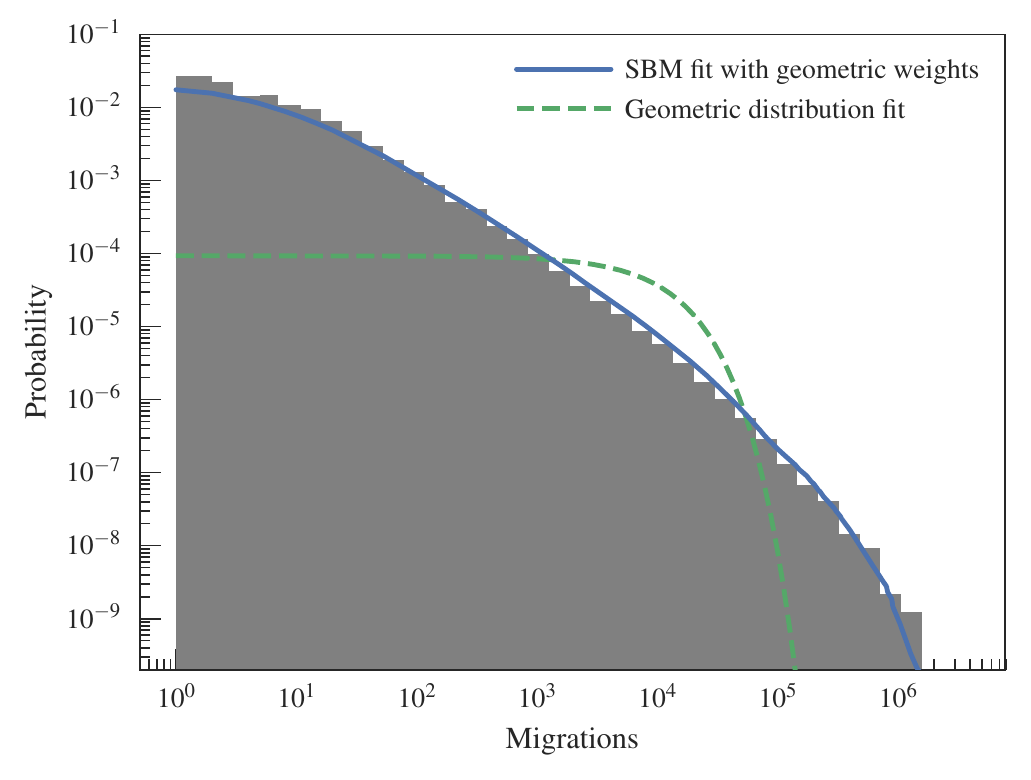}

  \caption{Overall distribution of the number of migrations for the UN
  data. The solid line shows the inferred distribution according to the
  weighted SBM using geometric distributions. The dashed line shows the
  best fit of a single geometric
  distribution.\label{fig:migrations-fit}}
\end{figure}
\begin{figure*}
  \begin{tabular}{cccc}
    \begin{overpic}[width=.33\textwidth]{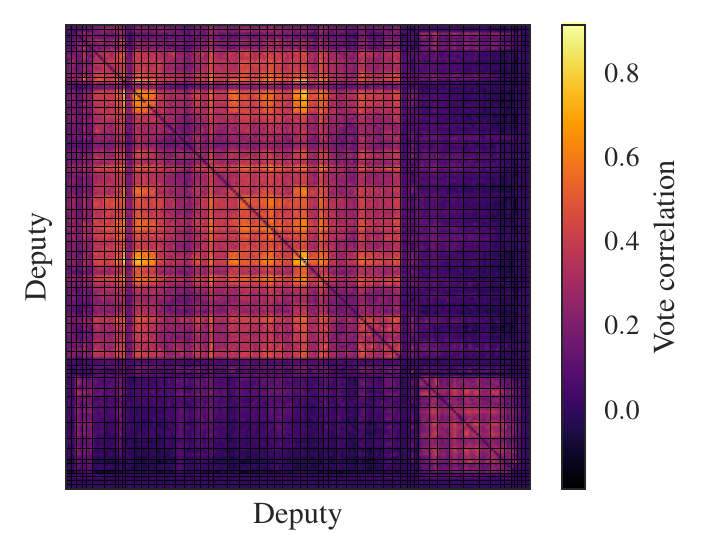}\put(0,0){(a)}\end{overpic}&
    \begin{overpic}[width=.26\textwidth,trim=0 -1cm 0 0]{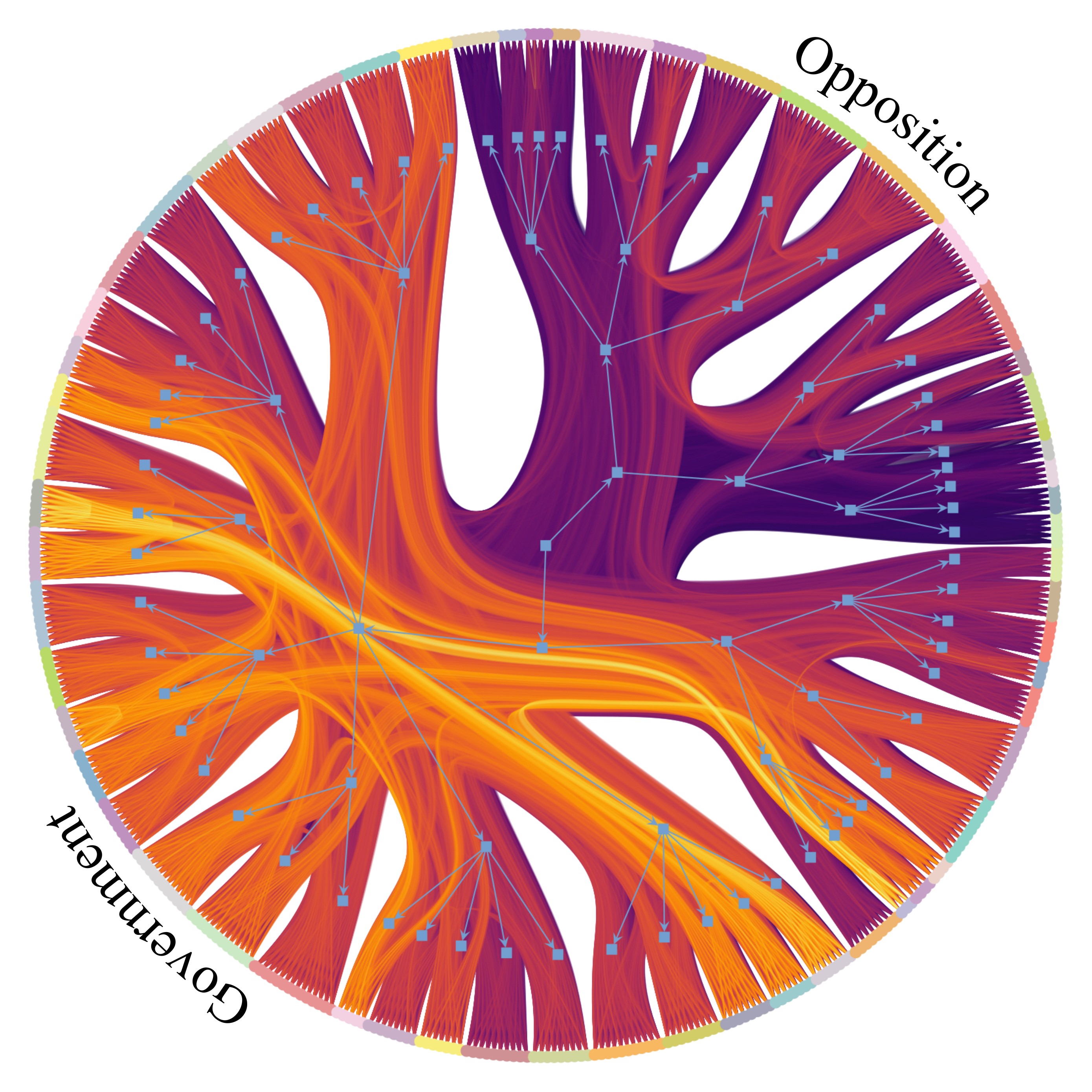}\put(0,0){(b)}\end{overpic}&
    \begin{overpic}[width=.32\textwidth]{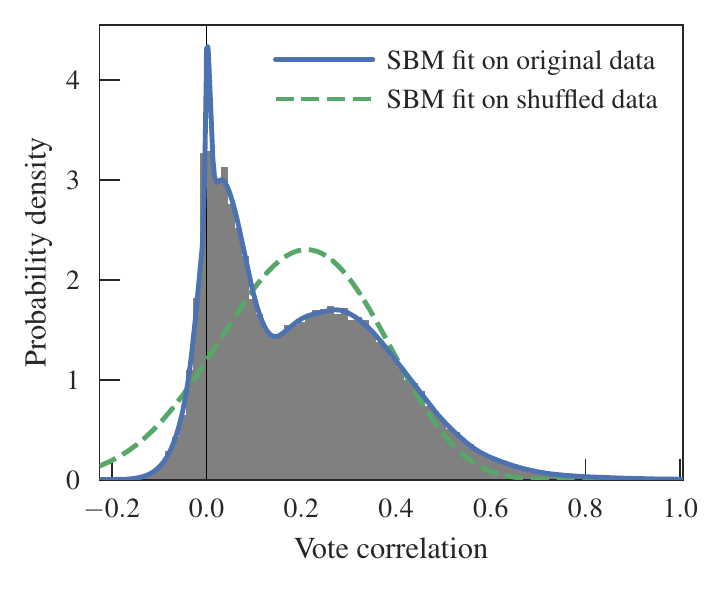}\put(0,0){(c)}\end{overpic}
    \end{tabular}
  \caption{(a) Fit of the weighted SBM for a matrix of vote correlations
  between deputies of the lower house of the Brazilian congress. The
  group boundaries are shown by horizontal and vertical lines. (b) Same
  as in (a) but using the layout of Fig.~\ref{fig:migrations} that shows
  the entire hierarchical division. (c) Overall distribution of vote
  correlations. The solid line shows the inferred distribution according
  to the weighted SBM using transformed normal distributions. The dashed
  line shows the best fit using the same model, but on the shuffled data
  with the same empirical distribution.\label{fig:votes}}
\end{figure*}

We begin with an illustration of how incorporating edge weights with our
method can have a significant effect on the analysis of network data. We
use for this purpose a dataset of global migrations between $N=232$
countries, assembled in 2015 by the United Nations~\footnote{Data
available at
\url{https://www.un.org/en/development/desa/population/migration/data/estimates2/estimates15.shtml}}. This
dataset can be represented as a directed network (see
Appendix~\ref{app:directed}), where for a pair of countries $(i,j)$
there is a net migrant stock $x_{ij} \in \mathbb{Z}$ which is defined as
the number of migrants that moved from $i$ to $j$ minus the number that
moved from $j$ to $i$. If we only had an unweighted SBM at our disposal,
a common approach would be to threshold this data, yielding a directed
edge $A_{ij} = 1$ if $x_{ij}>0$ and $A_{ij}=0$ otherwise. As argued by
Aicher et al~\cite{aicher_learning_2015}, this type of data manipulation
should be avoided whenever possible, since not only it destroys
potentially valuable information, but also it is possible to construct
examples where no single threshold can accurately reproduce the
large-scale structure in the data. In this particular case, this
approach actually does seem to yield usable information at first, as can
be seen in Fig.~\ref{fig:migrations}a, which shows a fit of the
unweighted SBM. We can see that the network division obtained in this
manner essentially categorizes countries on whether they are net sources
or targets of migration, as well as the typical regions people migrate
to and from. However, a closer inspection reveals that it is not able to
distinguish between countries like Costa Rica, South Africa and Finland
(which end up clustered in the same group as Austria and Ireland), which
not only are geographically far apart, but do have, in fact, very
distinct migration volumes and patterns. Since migration volumes between
countries can vary by several orders of magnitude (see
Fig.~\ref{fig:migrations-fit}), any analysis that ignores this aspect
must be woefully incomplete. Indeed, if we include the values $x_{ij}$
of the migrant stock, in addition to the same adjacency matrix obtained
with the threshold approach, and we use the weighted SBM defined
previously, with geometric distributions for the weights as described in
Sec~\ref{sec:geometric}, we obtain a much more detailed representation
of the data, as shown in Fig.~\ref{fig:migrations}b. Not only we find a
larger number of groups, but now countries like France, Canada and
United Kingdom appear as members of very specific groups. The United
States of America gets placed in its own group, due its unique volume
and pattern of (mostly incoming) migrations. The remaining countries end
up divided in geographically meaningful categories, with regions like
South America, Middle East, Africa and Asia being easily
recognizable. However, there are exceptions to this, where
geographically separated countries get clustered together. Examples of
this include Germany and India, as well as China and South Africa. These
countries are either sources or targets of global migration which goes
well beyond their immediate neighborhoods, and they possess similar
overall patterns despite geographical distance (we emphasize that due to
the degree-corrected nature of our model, countries with distinct
migration balances can be put in the same category, if their group
affinities are the same).

We can also assess the quality of the SBM in capturing the overall
weight distribution, computed from the model as
\begin{equation}
  P(x|\A,\x,\{\bb^l\}) = \frac{1}{E}\sum_{r\le s}m_{rs}^1P(x|\bar{\bm{\gamma}}_{rs}^1),
\end{equation}
with $E=\sum_{r\le s}m_{rs}^1$ being the total number of edges, and
$P(x|\bar{\bm{\gamma}}_{rs}^1)$ is the marginal covariate distribution between
groups $r$ and $s$, which in this case is given by
Eq.~\ref{eq:geom_marginal}. From Fig.~\ref{fig:migrations-fit}, we
see that the overall inferred distribution --- which is a particular
mixture of geometric distributions --- is capable of providing a very
good fit of the empirical data, despite the fact it is much broader than
any single geometric distribution (a best fit of which is shown for
reference).

\subsection{Vote correlations in congress}

We move now to another example where methods for unweighted graphs are
ill suited. We consider the voting patterns of $N=475$ members of the
lower house of the Brazilian congress during 2009~\footnote{Data
available from the official website \url{http://www2.camara.leg.br/}.}:
Each deputy voted ``yes'' or ``no'' on proposed laws during the
legislative year, and based on this, we computed the normalized
correlation between the votes $x_{ij}\in [-1,1]$ of deputies $i$ and
$j$. Note that in this case we have an adjacency matrix which is a
complete graph, i.e. $A_{ij}=1$ for all $i,j$, and any pairs with zero
correlation are considered particular values of the covariates.

This time we skip any attempt at thresholding the data, and we move
directly to the analysis using the weighted SBM. For this, we use the
version with normal distributions described in Sec.~\ref{sec:normal},
adapted to bounded weights via the variable transformation $y_{ij} =
2\operatorname{arctanh}(x_{ij})$ that maps the intervals
$[-1,1]\to[-\infty,\infty]$, as described in
Sec.~\ref{sec:transformations}. As shown in Fig.~\ref{fig:votes}a, the
method uncovers many groups of deputies, which collectively can be
divided into two overall groups at the highest hierarchical level. These
two large groups are more correlated with their own members than with
non-members. An inspection of the known party affiliations of the
deputies reveals that these two overall groups correspond to the
government and opposition, which tend to vote together either against or
in favor of bills. If we again inspect the overall distribution of vote
correlations, we see that the weighted SBM provides a very good fit, as
seen in Fig.~\ref{fig:votes}c. The model captures the bimodal nature of
the vote correlations --- with higher values corresponding to pairs of
deputies belonging both to either the government or opposition, and
lower values to pairs belonging to different factions. It should be
emphasized that the quality of the fit is not merely an outcome of using
a sufficiently large mixture of normal distributions, as we are not
modelling the overall distribution directly. Instead, the distributions
are tied to the division of the nodes into groups, and the quality of
the overall fit shows that the distribution of weights is well
correlated with this categorization. For comparison, we show in
Fig.~\ref{fig:votes}c the outcome of the same analysis where the exact
same weights are used, but they are randomly shuffled across pairs of
deputies, thereby destroying any group organization but preserving the
overall weight distribution. In this case, the best SBM fit is composed
of only one group, $B=1$, and the corresponding normal fit cannot
capture the bimodal structure of the weights --- although it is still
present in the shuffled data, albeit in a manner which is completely
uncorrelated with any partition of the deputies. Therefore a close match
between the empirical weight distribution and the SBM fit like the one
in Fig.~\ref{fig:votes}c --- as well as the one in
Fig.~\ref{fig:migrations-fit} for the UN migration data --- is a
testament to the quality of the SBM ansatz in explaining the data,
rather than of an arbitrary mix of elementary unimodal distributions.

\begin{figure*}
  \begin{tabular}{ccc}
    \begin{overpic}[width=.60\textwidth, trim=0 -.5cm 0cm 0]{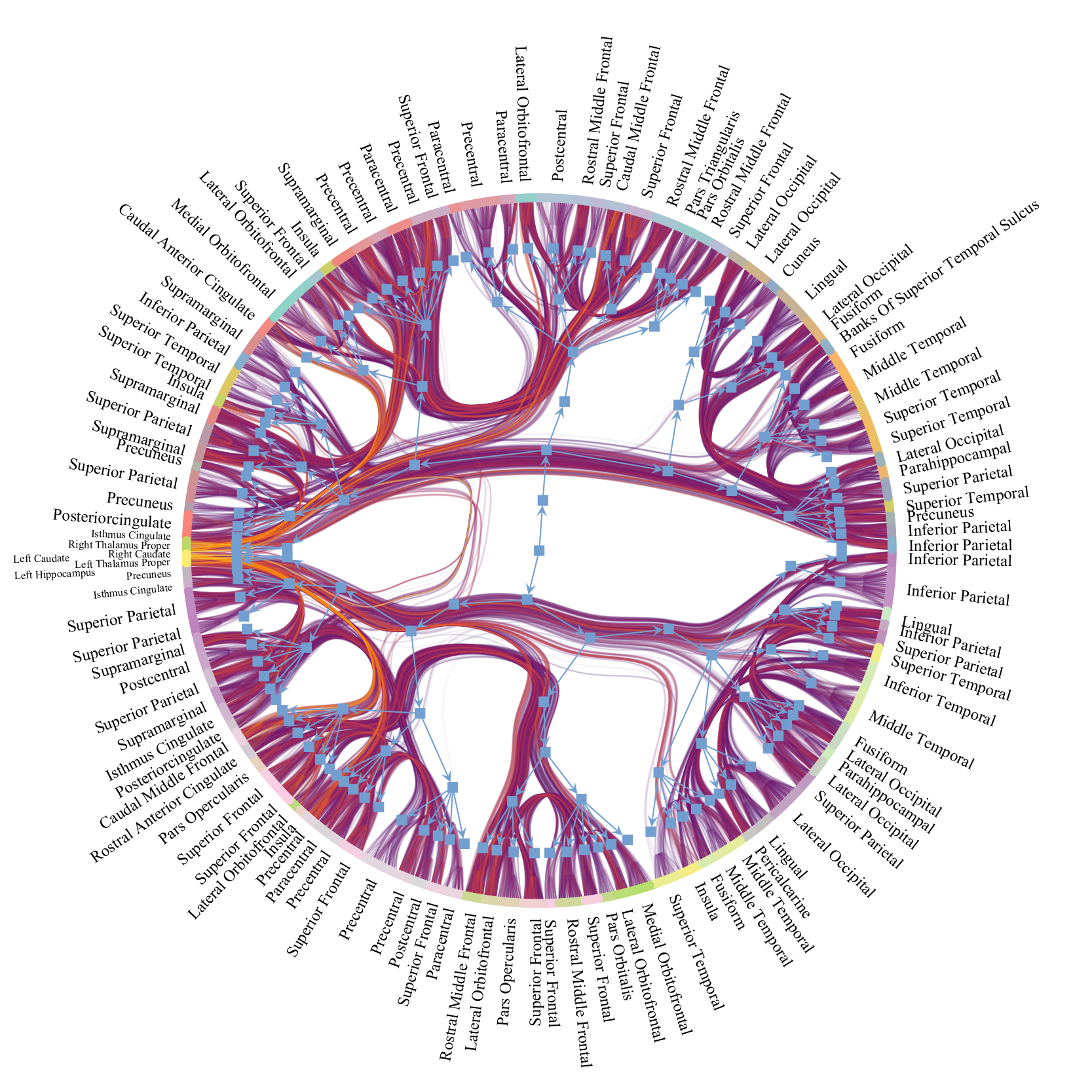}
      \put(0,0){(a)}
      \put(40, 98){\smaller Right hemisphere}
      \put(37, 0) {\smaller Left hemisphere}
    \end{overpic}&
        \raisebox{1.4cm}{\includegraphics[width=.38\textwidth,angle=90,origin=l]{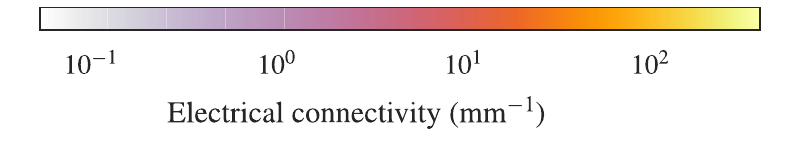}}&
    \raisebox{4.5cm}{
    \begin{tabular}{c}
      \begin{overpic}[width=.32\textwidth]{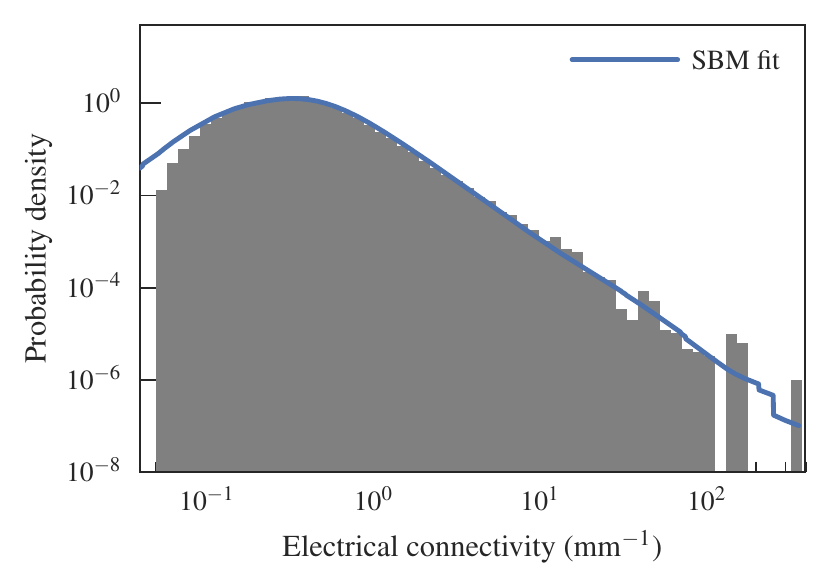}\put(0,0){(b)}\end{overpic}\\
      \begin{overpic}[width=.32\textwidth]{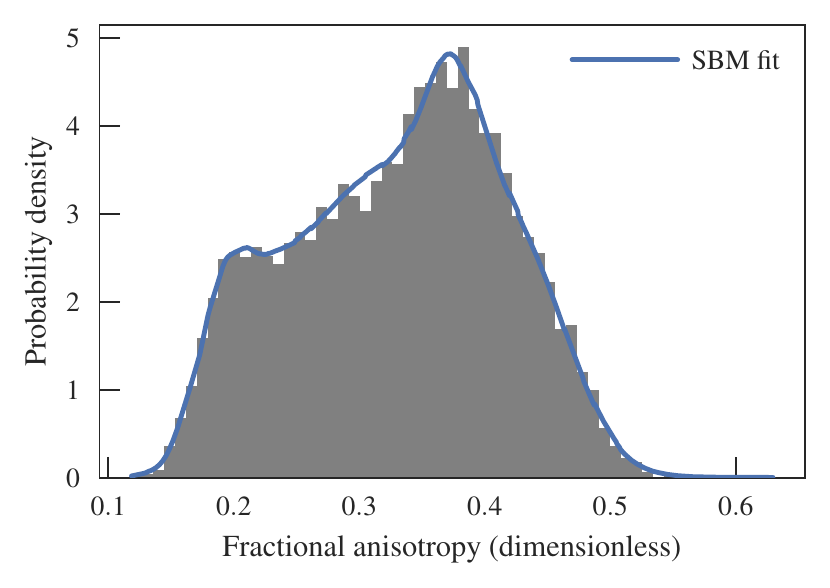}\put(0,0){(c)}\end{overpic}
    \end{tabular}}
  \end{tabular}

  \caption{(a) Inferred SBM for the human connectome, using electrical
  connectivity and fractional anisotropy as edge covariates. The text
  labels show the most frequent anatomical annotation inside each group
  at the lowest hierarchical level; (b) Empirical and fitted
  distribution of electrical connectivity of the edges; (c) Empirical
  and fitted distribution of fractional anisotropy of the
  edges. \label{fig:brain}}
\end{figure*}

\begin{figure*}
  \begin{tabular}{c}
    \begin{overpic}[width=\textwidth]{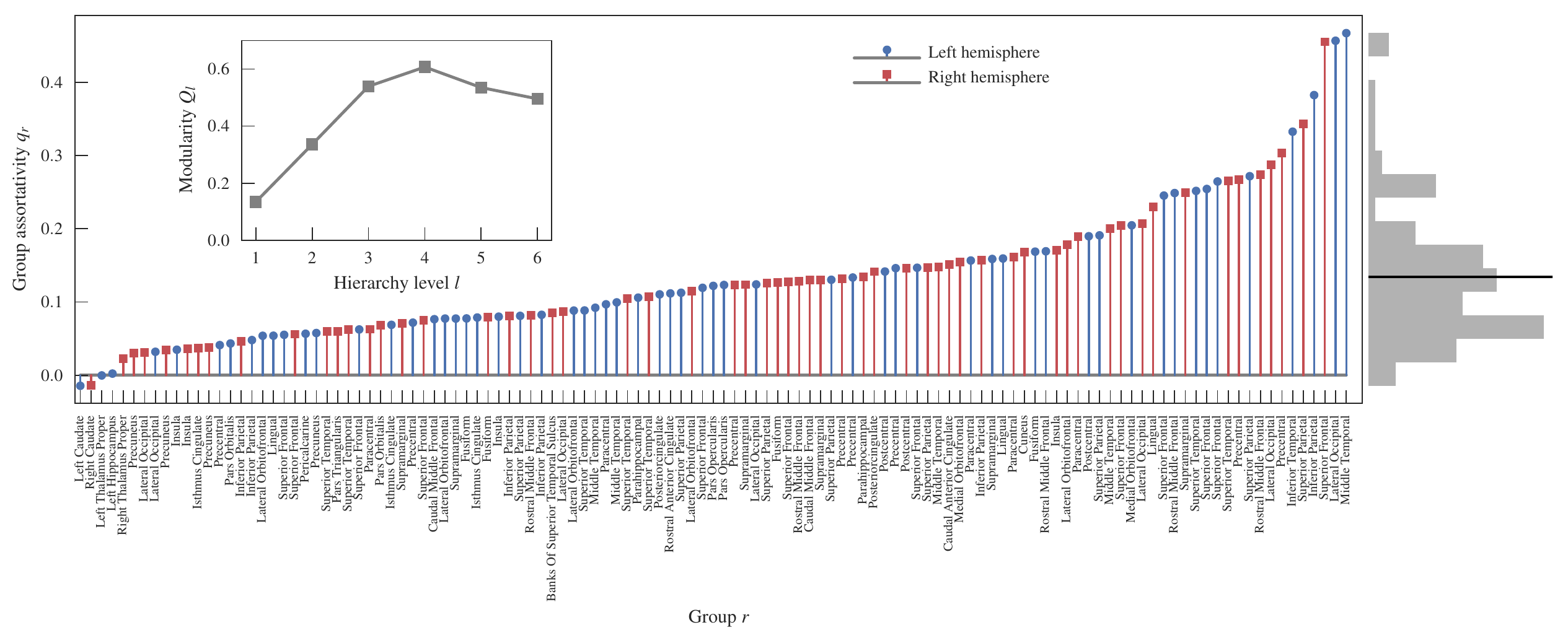}\put(0,1){(a)}\end{overpic}\\
    \begin{overpic}[width=\textwidth]{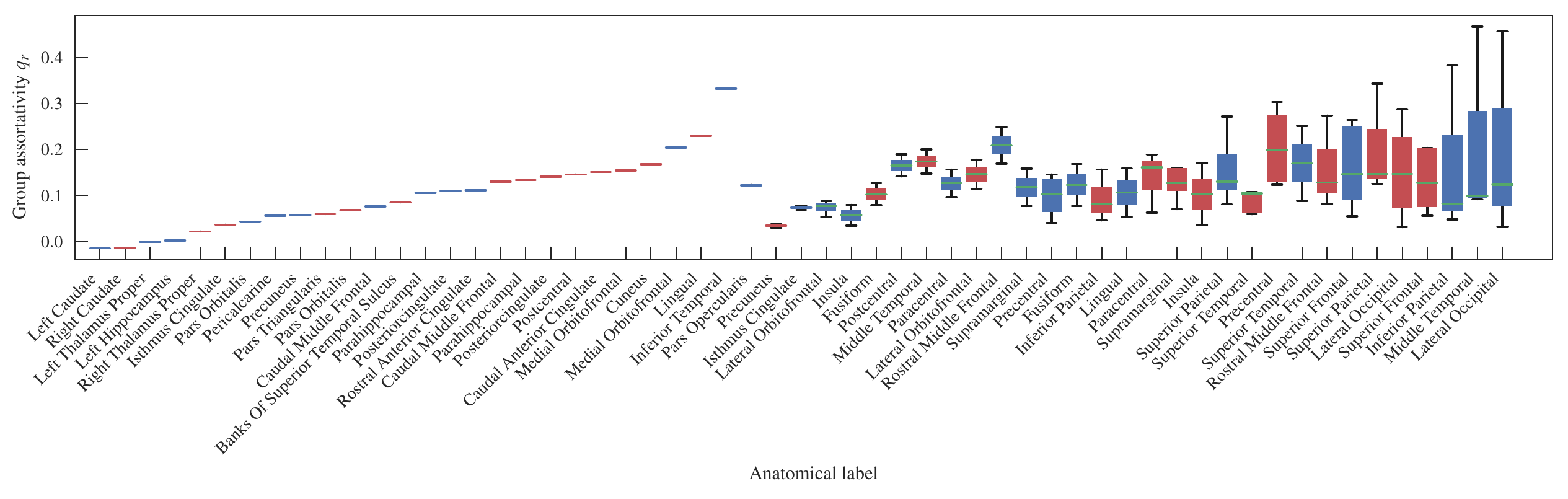}\put(0,1){(b)}\end{overpic}\\
  \end{tabular} \caption{(a) Group assortativity $q_r$ (Eq.~\ref{eq:qr})
  for the lowest level of the hierarchy in Fig~\ref{fig:brain}a, with
  groups labelled using the most frequent anatomical annotation. Blue
  circle (red square) markers correspond to the left (right)
  hemisphere. On the right axis is shown a histogram of the $q_r$
  values, with a horizontal line marking the average
  $Q=\sum_rq_r/B\approx 0.13$. The inset shows the modularity value
  $Q_l$ as a function of the hierarchy level $l$. (b) Dispersion of
  $q_r$ values for groups that share the same anatomical annotation, as
  labelled in the x-axis.\label{fig:brain-modularity}}
\end{figure*}

\subsection{The human brain}\label{sec:brain}

We now analyze empirical networks of interactions between parts of human
the brain, using data from the Budapest Reference
Connectome~\cite{szalkai_parameterizable_2017} (which itself is based on
primary data from the Human Connectome
Project~\cite{mcnab_human_2013}). This dataset corresponds to a
consensus between 477 people, where an edge between two of $N=1,006$
pre-defined anatomical regions is considered to exist, i.e. $A_{ij}=1$,
if neuronal fibers connecting these two regions have been detected in at
least 20\% of the individuals. In addition to this basic connectivity,
we consider two edge covariates, averaged over individuals: The
``electrical connectivity'' $x_{ij} \in [0,\infty]$, defined as the
number of recorded fibers divided by their length, and the fractional
anisotropy~\cite{basser_microstructural_1996}, $y_{ij} \in [0,1]$, which
is maximal if all fibers in the affected region go in the same direction
in 3D space, or minimal if they all go in different directions.  Indeed,
we use this dataset as an opportunity to highlight that our method can
also be used when there are multiple covariates available. This can be
done in an intuitive manner by assuming that their generation is
conditioned on the same network partition, but otherwise are
independent, i.e.
\begin{equation}
  P(\bm{x},\bm{y}|\A,\{\bb^l\})=P(\bm{x}|\A,\{\bb^l\})P(\bm{y}|\A,\{\bb^l\}).
\end{equation}
We can then use the exact same algorithm to obtain the posterior
$P(\{\bb^l\}|\A,\bm{x},\bm{y})$ by simply combining both terms for
$\bm{x}$ and $\bm{y}$. This approach will use the information in both
covariates simultaneously to inform the partition of the network. (This
is easily extended for an arbitrary number of covariates, and hence
yields a method that is also suitable for vector-valued covariates,
which is supported in our reference
implementation~\cite{peixoto_graph-tool_2014}.) In the following, we use
normal models for the transformed covariates $\ln x_{ij}$ and
$\operatorname{logit}(y_{ij})$.

When applied to the brain dataset, our method reveals the structure
shown in Fig.~\ref{fig:brain}a. It decomposes the network into left and
right hemispheres at the topmost hierarchical level, and proceeds to
subdivide it into smaller regions. The subdivisions in both hemispheres
are similar but not quite identical, indicating imperfect bilateral
symmetry. The divisions at the bottom level are well correlated with
known anatomical divisions, as shown by the labels in
Fig.~\ref{fig:brain}a. Most often, our method finds \emph{subdivisions}
of anatomical regions --- i.e. a single anatomical region is divided in
one or more groups --- which are then grouped together higher in the
hierarchy. But we also find some regions that belong to the same
anatomical group that end up classified in significantly different
hierarchy branches, pointing to a further degree of heterogeneity inside
anatomical regions. Since the various traditional approaches to
determine such anatomical classification do not always take into account
the local and global connectivity patterns (i.e. the actual connectome),
our approach suggests an alternative or complementary method to perform
such a task.

Like in the previous examples, the fit of the overall distributions of
edge covariates provided by the SBM is reasonably convincing, as we see
in Figs.~\ref{fig:brain}b and c, indicating that these nontrivial
distributions --- which deviate significantly from the basic
distributions used in the model --- can be well explained by
group-to-group mixtures.

\subsubsection{Community structure?}

The modular structure of the brain has been studied numerous times
before, using a variety of methods
(e.g.~\cite{meunier_modular_2010,bassett_robust_2013,betzel_multi-scale_2013}). Most
often, however, this is done by searching for \emph{assortative}
modules~\cite{newman_mixing_2003},
i.e. groups of nodes more connected to themselves than with the rest of
the network --- a pattern commonly called \emph{community
structure}~\cite{girvan_community_2002,fortunato_community_2010}. In
contrast, the approach developed here seeks to find groups of nodes that
have similar probabilities of connection with the rest of the network
(and to generate edge covariates), regardless if they form a community
or not. Naturally, community structure is a special case of the general
class of patterns that we consider, but our approach is capable of
accommodating many others, such as core-peripheries and bipartiteness
--- in fact, any arbitrary kind of group affinities. This means that if
the formation of assortative communities is the main driving mechanism
responsible for the network structure, we should be able to detect it
with our method, but otherwise it will prefer a non-assortative
division. This makes it a more flexible and potentially more informative
approach in comparison to typical community detection methods, which, by
construction, will tend to omit non-assortative divisions, however
important they may be, in favor of assortative ones. In the case of
brain networks, very often the community detection approach used is the
maximization of modularity~\cite{newman_mixing_2003}, defined as
\begin{equation}\label{eq:modularity}
  Q = \frac{1}{2E}\sum_r e_{rr} - \frac{e_r^2}{2E},
\end{equation}
where $e_{rs}=m_{rs}(1+\delta_{rs})$ is the number of edges between
groups $r$ and $s$ (or twice that if $r=s$), and $e_r=\sum_se_{rs}$. As
has been known for a long time~\cite{guimera_modularity_2004}, and since
then has become well
understood~\cite{reichardt_when_2006,bagrow_communities_2012,
  mcdiarmid_modularity_2013,mcdiarmid_modularity_2016,prokhorenkova_modularity_2016},
the direct maximization of $Q$ to detect communities will generically
\emph{overfit}, as it will misleadingly find many spurious communities
and produce large $Q$ values for completely random networks, as well as
arguably non-modular networks such as trees. Somewhat paradoxically, the
same approach will also generically \emph{underfit}, as it is incapable
of detecting a number of communities larger than
$\sqrt{2E}$~\cite{fortunato_resolution_2007}, even if their presence is
statistically significant. Because of these and other limitations, as
well as its non-statistical nature, the unsupervised maximization of $Q$
to find communities is ill-advised in most
contexts~\cite{fortunato_community_2016}. In contrast, the approach
presented here is free of both these problems: When applied to
completely random networks, it will not uncover spurious groups not
sufficiently backed by statistical
evidence~\cite{peixoto_parsimonious_2013}; and it is capable of
detecting up to $\propto N/\log N$
groups~\cite{peixoto_hierarchical_2014,peixoto_nonparametric_2017},
whenever they are present. Since we have principled guarantees that the
modules uncovered with our method are statistically significant, we can
then use the value of $Q$ to characterize the degree of assortativity of
the modules found (rather than the quality of the partition). For the
result shown in Fig.~\ref{fig:brain}, we obtain $Q\approx 0.13$, which
is typically considered a low value indicating weak community
structure. We may understand this value in more detail by decomposing it
as
\begin{equation}
  Q = \frac{1}{B}\sum_rq_r
\end{equation}
where
\begin{equation}\label{eq:qr}
  q_r = \frac{B}{2E}\left(e_{rr} - \frac{e_r^2}{2E}\right)
\end{equation}
is the local assortativity of group $r$, with $q_r\in[-1,1]$. In
Fig.~\ref{fig:brain-modularity}a we show the values of $q_r$ for the
modules inferred with our method, labelled according to most prominent
anatomical classification. We see that while most values are positive,
$q_r>0$, strictly indicating a degree of assortativity, they are
distributed across a broad range --- with regions like the \emph{Caudate
nucleus} (associated with motor functions) even showing dissortativity
with $q_r<0$ --- indicating that assortativity, although it is present,
is not an overwhelmingly dominant descriptor of the large-scale
structure (a similar point has been made
recently~\cite{betzel_diversity_2017} using the method of
Ref.~\cite{aicher_learning_2015}). We note also that inferred groups
that are associated with the same anatomical region sometimes possess
very different assortativity, as shown in
Fig.~\ref{fig:brain-modularity}b. This gives us an insight as to why
they were classified in different groups in the first place, and further
corroborating the idea that specific anatomical regions have noticeable
internal heterogeneity.

One might speculate that assortativity is just one of a diverse set of
driving forces behind the network formation, and that inspecting a
detailed model of the network might dilute its importance. Here we can
can further exploit the multilevel nature of our inferred model to
assess if assortativity becomes more relevant at higher levels of
coarse-graining. We can do so by computing a different value $Q_l$ for
each hierarchical level $l$, defined by replacing $m_{rs}\to m_{rs}^l$
in Eq.~\ref{eq:modularity}, where $m_{rs}^l$ is the number of edges
between groups $r$ and $s$ at level $l$. As we can see in the inset of
Fig.~\ref{fig:brain-modularity}a, the values of $Q_l$ do significantly
increase at higher levels, suggesting that assortativity might be an
important mechanism for the most global structures of the network, but
not as much for its sub-structures at a smaller scale.

\section{Elementary models for edge weights}\label{sec:likelihoods}

The central piece of the posterior distribution of hierarchical
partitions of Eq.~\ref{eq:posterior_b_l}
is the joint marginal probability of the network adjacency and weights,
\begin{equation}\label{eq:joint_marginal_b_l}
  P(\A,\x|\{\bb^l\}) =  P(\x|\A,\{\bb^l\}) P(\A|\{\bb^l\}).
\end{equation}
For the unweighted part, $P(\A|\{\bb^l\})$, we use the family of
unweighted nested SBMs developed in
Ref.~\cite{peixoto_hierarchical_2014, peixoto_nonparametric_2017}. The
reader is referred to those references, as well as the more recent
overview provided in Ref.~\cite{peixoto_bayesian_2017}, for a detailed
derivation of the unweighted marginal likelihood, which we omit here for
conciseness. To complete the model, we need to determine the placement
of edge weights given the adjacency matrix and the hierarchical
partition, with probability $P(\x|\A,\{\bb^l\})$ given by
Eq.~\ref{eq:marg_w_nested}, which depends on the nature of the edge
covariates.

In this section we derive models for edge covariates based on basic
properties, such as whether they are signed or unsigned, continuous or
discrete, bounded or unbounded. In particular, we focus on formulations
that allow the integrated marginal likelihood $P(\x|\A,\{\bb^l\})$ to be
computed exactly. For some of the derivations, we will assume --- for
convenience of notation --- that the graphs are simple,
i.e. $A_{ij}\in\{0,1\}$. We do so without loss of generality, as the
final expressions will also be valid for multigraphs. In all cases, we
begin with the simpler case of only one hierarchical level, where
$\{\bb^l\}$ is replaced by a single node partition $\bb$, and generalize
thereafter.

\subsection{Continuous unsigned weights}\label{sec:expon}

If all we know about the edge weights is that they are continuous and
positive, i.e. $x_{ij} > 0$, a reasonable model is a maximum-entropy
distribution with a fixed average, i.e. the exponential distribution
\begin{equation}\label{eq:expon}
  P(x|\lambda) = \lambda \ee^{-\lambda x}.
\end{equation}
Using this as the basis of our weighted SBM yields,
\begin{align}
  P(\x_{rs} | \A, \bm\lambda, \bb) &=
  \prod_{ij}P(x_{ij}|\lambda_{rs})^{\frac{A_{ij}\delta_{b_i,r}\delta_{b_i,s}}{1+\delta_{rs}}} \\
  &= \lambda_{rs}^{m_{rs}}\ee^{-\lambda_{rs}\mu_{rs}},
\end{align}
with
\begin{equation}\label{eq:mu}
  \mu_{rs}=\sum_{ij}\frac{A_{ij}x_{ij}\delta_{b_i,r}\delta_{b_j,s}}{1+\delta_{rs}}
\end{equation}
being the sum of the weights between groups $r$ and $s$.  Before
computing the integrated marginal likelihood of Eq.~\ref{eq:marg_w}, we
need to select a prior for $\bm{\lambda}$. A natural choice that makes
the computation feasible is known as a conjugate prior, which in this
case is the gamma distribution
\begin{equation}\label{eq:gamma}
  P(\lambda|\alpha,\beta) = \frac{\beta^{\alpha}\lambda^{\alpha-1}}{\Gamma(\alpha)}\ee^{-\lambda\beta},
\end{equation}
where $\alpha$ and $\beta$ are hyperparameters controlling its
shape. Using this prior, we can write the marginal likelihood for the
network weights by integrating over all $\lambda_{rs}$, yielding
\begin{equation}
  P(\x | \A, \bb, \alpha, \beta) = \prod_{r\le s}\frac{\Gamma(m_{rs}+\alpha)}{\Gamma(\alpha)}\frac{\beta^{\alpha}}{(\mu_{rs}+\beta)^{m_{rs}+\alpha}}.
\end{equation}
Doing so, we have reduced the initially high number of parameters from
$B(B+1)/2$ to only two, corresponding to the hyperparameters $\alpha$
and $\beta$. Being global parameters, independent of the internal
dimension of the model, they can be chosen via maximum likelihood,
without significant risk of overfitting,
\begin{equation}
   \hat{\alpha}, \hat{\beta} = \underset{\alpha,\beta}{\operatorname{argmax}}\; P(\x | \A, \bb, \alpha, \beta),
\end{equation}
which can be done efficiently with any standard optimization
method. Alternatively, we may consider the choice $\alpha=1$, for which
$P(\lambda|\alpha,\beta)$ becomes the maximum-entropy distribution with
a fixed mean, and hence has the same shape as $P(x|\lambda)$. Even with
this choice, however, is difficult to incorporate this prior in the
nested SBM via Eq.~\ref{eq:marg_w_nested}, as the integration over the
remaining hierarchical levels is cumbersome. Instead, we now describe a
microcanonical formulation which generates covariates in an
asymptotically identical manner, but permits the exact integration of
Eq.~\ref{eq:marg_w_nested}.

\subsubsection{Microcanonical distribution}

\begin{figure}[h!]
  \includegraphics[width=\columnwidth]{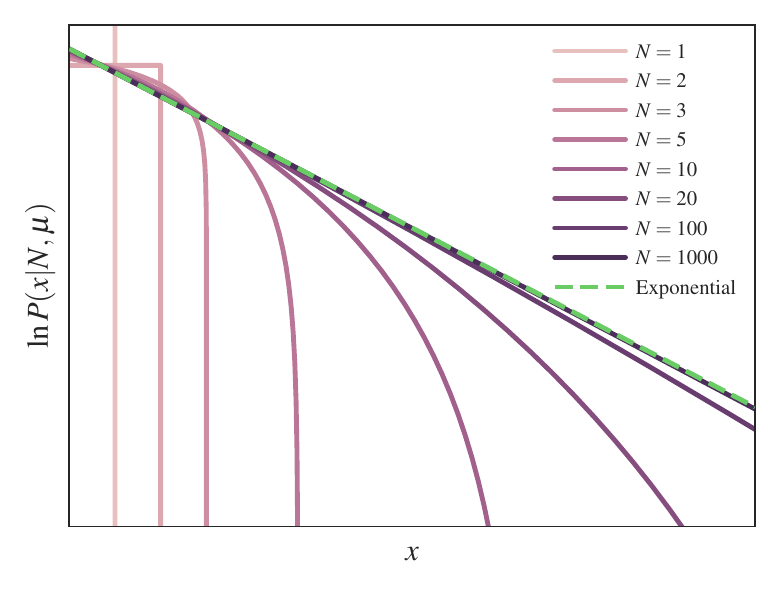} \caption{The
  marginal distribution of each individual covariate $x$ in the unsigned
  microcanonical model, given by Eq.~\ref{eq:unsigned_marginal},
  approaches asymptotically the exponential distribution as the number
  of values $N$ increases, and if the mean $\bar{x}=\mu/N$ is kept
  fixed.\label{fig:expon}}
\end{figure}

Instead of generating each covariate independently, we consider the
uniform joint distribution of $N$ positive real values
$\x=\{x_1,\dots,x_N\}$ conditioned on their total sum $\mu=\sum_ix_i$,
\begin{equation}\label{eq:expon_micro}
  P(\x | \mu) =
  \begin{cases}{\displaystyle
    \frac{(N-1)!}{\mu^{N-1}}}\delta(\mu-\sum_ix_i) & \text{ if } \mu > 0\\
    \prod_i\delta(x_i) & \text{ if } \mu = 0,
  \end{cases}
\end{equation}
where the normalization constant $(N-1)!/\mu^{N-1}$ above accounts for
the volume of a scaled simplex of dimension $N-1$. Although the
covariates are not generated independently in Eq.~\ref{eq:expon_micro},
the marginal distribution of the individual values $x_i$ can be
obtained as
\begin{align}
  P(x_i |N,\mu) &= \frac{P(\x|\mu)}{P(\x\setminus x_i|\mu-x_i)},\label{eq:expon_marg}\\
   &= \frac{(N-1)(\mu-x_i)^{N-2}}{\mu^{N-1}}\Theta(\mu-x_i)\label{eq:unsigned_marginal}
\end{align}
using Eq.~\ref{eq:expon_micro} both in the numerator and denominator of
Eq.~\ref{eq:expon_marg}, and $\Theta(x)$ is the Heaviside step
function. Taking the limit $N\to\infty$ while keeping the mean
$\bar{x}=\mu/N$ fixed, $P(x_i|N,\mu)$ becomes Eq.~\ref{eq:expon} with
$\lambda=1/\bar{x}$ (see Fig.~\ref{fig:expon}). Since in the limit of
sufficient data both models become identical, the microcanonical model
enables us to have an exact hierarchical SBM, as we will now show,
without sacrificing descriptive power.

Incorporating the microcanonical model of Eq.~\ref{eq:expon_micro} in the
SBM amounts simply to
\begin{equation}
  P(\x_{rs} | \A, \bm\lambda, \bb) = P(\x_{rs}|\mu_{rs}),\vspace{.5em}
\end{equation}
where, as before, $\mu_{rs}$ is the sum of covariates between groups $r$
and $s$. To generate the parameters $\mu_{rs}$ --- which are also
non-negative real numbers --- we can use the exact same distribution
again at a higher hierarchical level, by treating them as edge
covariates of the graph of groups, as described in
Eq.~\ref{eq:marg_w_nested}. The microcanonical nature of this model
makes the integration over all parameters $\{\bm\mu_{rs}^l\}$
trivial due to the hard constraints, i.e.
\begin{widetext}
\begin{align}
  P(\x | \A,\{\bb^l\})
  &= \int P(\x | \A, \bm{\mu}^1, \bb^1)\prod_{l=1}^L\prod_{r\le s}\left[P(\bm\mu^l_{rs}|\mu^{l+1}_{b_r^{(l+1)},b_s^{(l+1)}})\;\dd\bm\mu^l_{rs}\right]^{1-\delta_{m_{rs}^l,0}}\\
  &= \prod_{l=1}^L\prod_{r\le s}\left[\frac{(m_{rs}^l-1)!}{(\bar\mu_{rs}^l)^{m_{rs}^l-1}}\right]^{1-\delta_{\mu_{rs}^l,0}},\label{eq:marg_w_nested_unsigned}
\end{align}
\end{widetext}
where
\begin{equation}\label{eq:bar_mu}
  \bar\mu_{rs}^l=\sum_{tu}\left(\frac{\bar\mu_{tu}^{l-1}\delta_{b_t^l,r}\delta_{b_u^l,s}}{1+\delta_{rs}}\right)^{1-\delta_{m_{tu}^l,0}}
\end{equation}
is the sum of covariates between groups $r$ and $s$ at level $l>1$, and
with $\bar\mu_{rs}^1=\mu_{rs}$ given by Eq.~\ref{eq:mu} at the lowest
level. Recall that the boundary condition used in
Eq.~\ref{eq:marg_w_nested} is that at the topmost level there is only
one group, and hence $m_{rs}^L=E\delta_{r,1}\delta_{s,1}$ and
$\bar\mu_{rs}^L=\hat{\mu}\delta_{r,1}\delta_{s,1}$, where
$\hat{\mu}=\sum_{i<j}A_{ij}x_{ij}$ is the total sum of edge weights, and
the sole remaining parameter of the model. The marginal likelihood of
Eq.~\ref{eq:marg_w_nested_unsigned} is a simple term that can be
computed easily by obtaining the covariate summaries at each level, and
amounts to a straightforward modification of the algorithm of
Ref.~\cite{peixoto_nonparametric_2017} to obtain the posterior
distribution of hierarchical partitions. In particular, this additional
term does not affect its algorithm complexity, since changes in a lower
hierarchical level that are compatible with the partition at higher
level do not alter the likelihoods in the upper levels, as the covariate
sums remain unchanged.

\subsection{Continuous signed weights}\label{sec:normal}

For weights that can be either positive or negative, we require a
maximum entropy distribution with fixed average and variance, which is
the normal distribution
\begin{equation}
  P(x|\bar{x}, \sigma^2) = \frac{1}{\sqrt{2\pi\sigma^2}} \ee^{-\frac{(x-\bar{x})^2}{2\sigma^2}}.
\end{equation}
Incorporating this in the SBM, we obtain
\begin{align}
  P(\x_{rs} | \A, \bar{x}_{rs}, \sigma^2_{rs})
  &= \prod_{ij}P(x_{ij}|\bar{x}_{rs}, \sigma^2_{rs})^{\frac{A_{ij}\delta_{b_i,r}\delta_{b_i,s}}{1+\delta_{rs}}} \\
  &= \frac{\ee^{-\frac{\nu_{rs}-2\mu_{rs}\bar{x}_{rs}+m_{rs}\bar{x}_{rs}^2}{2\sigma_{rs}^2}}}{(2\pi\sigma_{rs}^2)^{m_{rs}/2}},
\end{align}
with
\begin{equation}\label{eq:nu}
\nu_{rs} =
\sum_{ij}\frac{A_{ij}x_{ij}^2\delta_{b_i,r}\delta_{b_j,s}}{1+\delta_{rs}}
\end{equation}
being the sum of squares of covariates between groups $r$ and $s$.
The conjugate prior for $\bar{x}$ and $\sigma^2$ is the
normal-inverse-chi-squared distribution~\cite{box_bayesian_2011}
\begin{equation}
  P(\bar{x}, \sigma^2 | \mu_0,\kappa_0,\nu_0,\sigma^2_0) =
  \mathcal{N}(\bar{x}|\mu_0, \sigma^2/\kappa_0)
  \chi^{-2}(\sigma^2|\nu_0,\sigma^2_0),
\end{equation}
where $\mathcal{N}(\bar{x}|a,b)$ is a normal distribution with mean $a$
and variance $b$, and the variance is sampled from an
inverse-chi-squared distribution
\begin{equation}
  \chi^{-2}(\sigma^2|\upsilon,\tau^2) = \frac{(\tau^2\upsilon/2)^{\upsilon/2}}{\Gamma(\upsilon/2)}~
\frac{\ee^{\frac{-\upsilon \tau^2}{2 \sigma^2}}}{\sigma^{2+\upsilon}}.
\end{equation}
Using this prior, after the integration over $\bar{x}$ and $\sigma^2$,
the marginal likelihood becomes
\begin{multline}
  P(\x_{rs} | \A, \mu_0,\kappa_0,\nu_0,\sigma^2_0) =\\
  \frac{\Gamma(\nu_{rs}'/2)}{\Gamma(\nu_0/2)}\sqrt{\frac{\kappa_0}{\kappa_{rs}}}\frac{(\nu_0\sigma^2_0)^{\nu_0/2}}{(\nu_{rs}'S_{rs})^{\nu_{rs}/2}}\frac{1}{\pi^{m_{rs}/2}},
\end{multline}
with auxiliary quantities
\begin{align}
  \kappa_{rs} &= \kappa_0 + m_{rs},\quad
  \nu_{rs}' = \nu_0 + m_{rs},\\
  z_{rs} &= \nu_{rs}-\mu_{rs}^2/m_{rs},\label{eq:z_sub}\\
  S_{rs} &= \frac{1}{\nu_{rs}'}\left[\nu_0\sigma_0^2 + z_{rs} + \frac{m_{rs}\kappa_0}{\kappa_0+m_{rs}}\left(\mu_0 - \frac{\mu_{rs}}{m_{rs}}\right)^2\right].
\end{align}
This leaves us with four global parameters, $\mu_0$, $\kappa_0$, $\nu_0$
and $\sigma^2_0$, that we have to determine either with maximum
likelihood, or maximum entropy arguments. However, like the unsigned
case previously, the shape of the marginal likelihood leaves little
chance of building a hierarchical model in closed form. Luckily, we can
once more construct a microcanonical model that allows us do precisely
that.

\subsubsection{Microcanonical distribution}

\begin{figure}
  \includegraphics[width=\columnwidth]{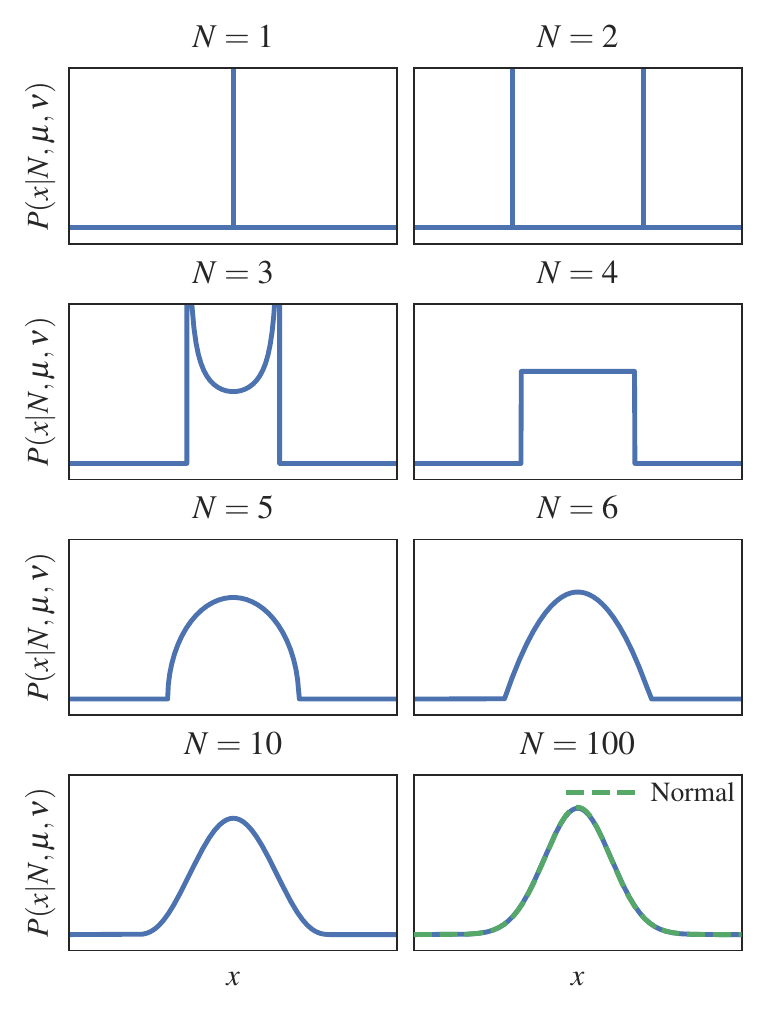}

  \caption{The marginal distribution of each individual covariate $x$ in
  the signed microcanonical model, given by
  Eq.~\ref{eq:signed_marginal}, approaches asymptotically the normal
  distribution as the number of values $N$ increases, and if the mean
  $\bar{x}=\mu/N$ and variance $\sigma^2=\nu/N-\bar{x}^2$ are kept
  fixed.\label{fig:normal}}
\end{figure}

The corresponding microcanonical maximum-entropy formulation for signed
covariates is the uniform distribution of $N$ values $\x$ conditioned in
the total sum $\mu$ and sum of squares $\nu$,
\begin{equation}\label{eq:normal}
  P(\x | \mu, \nu) = \frac{\delta(\mu-{\textstyle\sum_ix_i})\delta(\nu-{\textstyle\sum_ix_i^2})}{\Omega}.
\end{equation}
The normalization constant is computed as
\begin{align}
  \Omega&=\int \delta(\mu-{\textstyle\sum_ix_i})\delta(\nu-{\textstyle\sum_ix_i^2})\,\dd\bm{x}\\
  &=\int_H\frac{\delta(\textstyle\nu-\sum_ix_i(\bm{y})^2)}{\sqrt{N}}\,\dd\bm{y}(\bm{x})\label{eq:int_H}\\
  &=\int_S \frac{\dd\bm{\sigma}(\bm{x})}{2\sqrt{N\nu-\mu^2}}\label{eq:int_S}\\
  &=\frac{\pi^{(N-1)/2}\left(\nu-\mu^2/N\right)^{(N-3)/2}}{\Gamma(N/2-1/2)\sqrt{N}}\label{eq:signed_micro_C},
\end{align}
where $H$ in Eq.~\ref{eq:int_H} is the hyperplane given by
$\textstyle\sum_ix_i=\mu$, parametrized by $N-1$ coordinates
$\bm{y}(\bm{x})$, and $S$ in Eq.~\ref{eq:int_S} is the intersection of a
$N$-sphere of radius $\sqrt{\nu}$ and the hyperplane $H$, which
corresponds to the surface of a $(N-1)$-sphere of radius
$\sqrt{\nu-\mu^2/N}$, with surface element $\dd\bm{\sigma}(\bm{x})$,
leading to Eq.~\ref{eq:signed_micro_C}. Therefore, we have for the
complete microcanonical distribution
\begin{widetext}
\begin{equation}\label{eq:normal_micro}
  P(\x | \mu, \nu) =\\
  \begin{cases}\displaystyle
    \frac{\displaystyle\Gamma(N/2-1/2)\sqrt{N}}{\displaystyle\pi^{(N-1)/2}(\nu-\mu^2/N)^{(N-3)/2}}\delta(\mu-{\textstyle\sum_ix_i})\delta(\nu-{\textstyle\sum_ix_i^2}),&
    \text{ if } \nu > \mu^2/N, \\
    \prod_i\delta(\mu/N-x_i),& \text{ if } \nu = \mu^2/N.
    \end{cases}
\end{equation}
The marginal distribution of the individual covariates $x_i$ can be
obtained as
\begin{align}
  P(x_i |N,\mu,\nu) &= \frac{P(\x|\mu,\nu)}{P(\x\setminus x_i|\mu-x_i,\nu-x_i^2)},\label{eq:normal_marg}\\
   &= \frac{\Gamma(N/2-1/2)}{\Gamma(N/2-1)}\sqrt{\frac{N}{\pi(N-1)}}\frac{[\nu-x_i^2-(\mu-x_i)^2/(N-1)]^{N/2-2}}{(\nu-\mu^2/N)^{(N-3)/2}}\Theta(\mu-x_i)\Theta(\nu-x_i^2),\label{eq:signed_marginal}
\end{align}
\end{widetext}
using Eq.~\ref{eq:normal_micro} both in the numerator and denominator of
Eq.~\ref{eq:normal_marg}. Taking the limit $N\to\infty$ while keeping
both the mean $\bar{x}=\mu/N$ and variance $\sigma^2=\nu/N-\bar{x}^2$
fixed, $P(x_i|N,\mu, \nu)$ becomes the normal distribution of
Eq.~\ref{eq:normal} (see Fig.~\ref{fig:normal}). Therefore, like with the
unsigned case, the microcanonical model yields an easy-to-integrate
model, without sacrificing descriptive power.

Incorporating the above distribution into the SBM yields
\begin{equation}
  P(\x_{rs} | \A, \bm\mu, \bm\nu, \bb) = P(\x_{rs}|\mu_{rs},\nu_{rs}),
\end{equation}
where, as before, $\mu_{rs}$ is the sum of covariates between groups $r$
and $s$, and $\nu_{rs}$ is the sum of squares of the same covariates. To
generate the parameters $\mu_{rs}$, we can use the exact same
distribution again at a higher hierarchical level. The parameters
$\nu_{rs}$, however, are strictly positive, and hence require a
different model. Furthermore, $\mu_{rs}$ and $\nu_{rs}$ are not
independent parameters, as they must satisfy the inequality $\nu_{rs}
\ge \mu_{rs}^2/m_{rs}$. Therefore, we re-parametrize the model using the
auxiliary quantity of Eq.~\ref{eq:z_sub}
\begin{equation}
  z_{rs} = \nu_{rs} - \mu_{rs}^2/m_{rs},
\end{equation}
which is simply the scaled variance of the covariates, and thus is
strictly non-negative and can be chosen independently from
$\mu_{rs}$. We can then generate $z_{rs}$ from the unsigned
microcanonical model of Eq.~\ref{eq:expon_micro}. Although we can easily
write the final marginal likelihood of the model if we propagate the
hyperpriors of $z_{rs}$ upwards in the hierarchy of the nested SBM, we
would have the following problems: Not only this would increase the
total number of edge covariates at the highest levels (and hence it is
unclear \emph{a priori} if it is the most parsimonious approach), since
each signed parameter requires two hyperparameters, but also it leads to
a model that is cumbersome computationally, as changes in a lower level
would always propagate through the whole hierarchy. Instead, here we opt
to propagate only $\mu_{rs}$ upwards in the hierarchy, whereas we
generate all $z_{rs}$ from the same distribution at each level. More
concretely, we write
\begin{widetext}
\begin{align}
  P(\x | \A,\{\bb^l\})
  &= \int P(\x | \A, \bm{\mu}^1, \bm{z}^1, \bb^1)P(\bm\mu_z)\prod_{l=1}^LP(\bm z^l|\mu_z^l)\prod_{r\le s}\left[P(\bm\mu^l_{rs}|\mu^{l+1}_{b_r^{(l+1)},b_s^{(l+1)}},\bm{z}^{l+1}_{b_r^{(l+1)},b_s^{(l+1)}})\right]^{1-\delta_{m_{rs}^l,0}}\;\dd\bm\mu^l\dd\bm{z}^l\dd\mu_z^l,\label{eq:marg_w_nested_signed_pre}\\
  &=
  \frac{(\bar{L}-1)!}{(\sum_{l=1}^Lm_z^l\bar\mu_z^l)^{\bar{L}-1}}
  \prod_{l=1}^L(m_z^l)^{1-\delta_{m_z^l,0}}\left[\frac{(m_z^l-1)!}{(\bar\mu_z^l)^{m_z^l-1}}\right]^{1-\delta_{\bar\mu_z^l,0}}
  \prod_{r\le s}\left[\frac{\Gamma(m_{rs}^l/2-1/2)\sqrt{m_{rs}^l}}{\pi^{(m_{rs}^l-1)/2}(\bar{z}_{rs}^l)^{(m_{rs}^l-3)/2}}\right]^{1-\delta_{\bar{z}_{rs}^l,0}},\label{eq:marg_w_nested_signed}
\end{align}
\end{widetext}
where $\bar{z}_{rs}^l = \bar\nu_{rs}^l - (\bar\mu_{rs}^l)^2/m^l_{rs}$,
with $\bar\mu_{rs}^l$ given by Eq.~\ref{eq:bar_mu}, and
\begin{equation}\label{eq:bar_nu}
  \bar\nu_{rs}^l=\sum_{tu}\left[\frac{(\bar\mu_{tu}^{l-1})^2\delta_{b_t^l,r}\delta_{b_u^l,s}}{1+\delta_{rs}}\right]^{1-\delta_{m_{tu}^l,0}},
\end{equation}
corresponds to the scaled variance of the values of $\bar\mu_{rs}^{l-1}$
at a lower level (assuming the boundary conditions
$\bar\mu_{rs}^1=\mu_{rs}$ and $\bar\nu_{rs}^1=\nu_{rs}$ given by
Eqs.~\ref{eq:mu} and~\ref{eq:nu}, respectively), and where
\begin{align}
    m_z^l&=\sum_{r\le s}H(m_{rs}^l-1),\\
    \bar\mu_z^l&=\sum_{r\le s}\bar{z}_{rs}^lH(m_{rs}^l-1),
\end{align}
are the sum and scaled average $\bar{z}_{rs}^l$ of $m_{rs}^l>1$ entries
at level $l$, with $H(x)=1$ if $x>0$, otherwise $H(x)=0$, and $P(\bm
z^l|\mu_z^l)$ is given by Eq.~\ref{eq:expon_micro}. The above means that
when computing $P(\bm z^l|\mu_z^l)$ we must only consider values of
$z_{rs}^l$ for which $m_{rs}^l>1$. Otherwise, if $m_{rs}^l=1$, the
corresponding parameter must always be $\nu_{rs}^l=(\mu_{rs}^l)^2$, and
hence $z_{rs}^l=0$, which does not need to be sampled from a
prior. Finally, the values of $\bm\mu_z=\{\mu_z^l\}$ across all levels
are also sampled from their own model as
\begin{equation}
  P(\bm\mu_z) = P(\{m_z^l\mu_z^l\}|{\textstyle\sum_{l=1}^Lm_z^l\mu_z^l})\prod_{l=1}^L(m_z^l)^{1-\delta_{m_z^l,0}},
\end{equation}
using again Eq.~\ref{eq:expon_micro}, and where the trailing product is
a derivative term that accounts for the scaling of the variables in the
argument of the first term, and with
\begin{equation}
  \bar{L} = \sum_{l=1}^LH(m_z^l)
\end{equation}
being the number of levels with non-zero values of $m_z^l$. The boundary
condition in Eqs.~\ref{eq:marg_w_nested}
and~\ref{eq:marg_w_nested_signed_pre}, i.e.  that the last level of the
hierarchy has only one group, means that the two remaining parameters
are $\bar\mu_{rs}^L=\hat{\mu}\delta_{r,1}\delta_{s,1}$, where
$\hat{\mu}=\sum_{i<j}A_{ij}x_{ij}$ is the total sum of edge weights, and
$\hat{\mu}_z=\sum_lm_z^l\bar\mu_z^l$ which is the sum of scaled
variances across the hierarchy levels.

Like with the unsigned model, Eq.~\ref{eq:marg_w_nested_signed} amounts
to a straightforward modification of the algorithm of
Ref.~\cite{peixoto_nonparametric_2017}, requiring only an additional
book-keeping of the values of $z_{rs}^l$ for which $m_{rs}^l$ is larger
than one, and their respective sums, which can be done without altering
the overall algorithmic complexity. We remark also that, unlike maximum
likelihood approaches applied directly to Eq.~\ref{eq:normal}, the
resulting marginal likelihood of the microcanonical model is well
defined and yields non-degenerate results for any possible set of
covariates, even those yielding zero variance or populations with single
elements.

\subsection{Geometric discrete weights}\label{sec:geometric}

For discrete non-negative weights, i.e. $x\in\mathbb{N}_0$, the maximum
entropy distribution with a fixed average is the geometric distribution
\begin{equation}
  P(x|p) = (1-p)^xp.
\end{equation}
Using it for the SBM, we have
\begin{align}
  P(\x_{rs} | \A, \bb, p_{rs}) = (1-p_{rs})^{\mu_{rs}}p_{rs}^{m_{rs}}.
\end{align}
The conjugate prior for $p$ is the beta distribution
\begin{equation}\label{eq:beta}
  P(p|\alpha,\beta) = \frac{p^{\alpha-1}(1-p)^{\beta-1}}{B(\alpha,\beta)},
\end{equation}
with $B(x,y)=\Gamma(x)\Gamma(y)/\Gamma(x+y)$, which yields the marginal
distribution
\begin{align}
  P(\x_{rs} | \A, \bb, \alpha, \beta) = \frac{B(m_{rs}+\alpha, \mu_{rs}+\beta)}{B(\alpha,\beta)}.
\end{align}
Unlike the continuous case, we can make a fully ``uninformative'' choice
$\alpha=\beta=1$ that reflects our maximum ignorance about the parameter
$p$, as in this case it is uniformly sampled in the interval
$p\in[0,1]$. This yields simply
\begin{align}
  P(\x_{rs} | \A, \bb) = \frac{m_{rs}!\,\mu_{rs}!}{(m_{rs}+\mu_{rs}+1)!}.
\end{align}
However, this kind of uninformative assumption rarely matches what we
end up finding in the data, which tends to be significantly more
structured. A more robust approach is to construct a hierarchical model,
which can be more easily done with a microcanonical description.

\begin{figure}
  \includegraphics[width=\columnwidth]{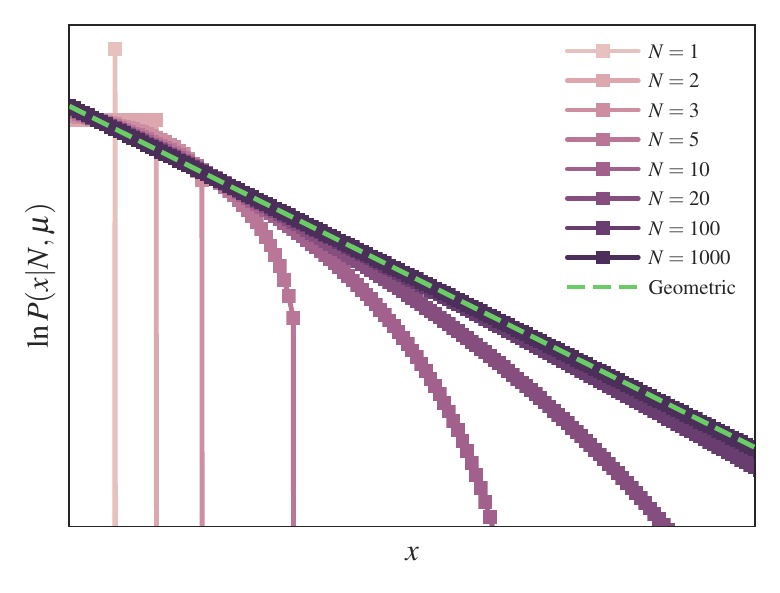}

  \caption{The marginal distribution of each individual covariate $x$ in
  the discrete microcanonical model, given by
  Eq.~\ref{eq:geom_marginal}, approaches asymptotically the geometric
  distribution as the number of values $N$ increases, and the mean value
  $\mu/N$ is kept fixed.\label{fig:geom}}
\end{figure}

\subsubsection{Microcanonical distribution}

The microcanonical analogue of the geometric distribution is the uniform
distribution of $N$ non-negative discrete real values $\x$ conditioned
on their total sum $\mu$, given by
\begin{equation}\label{eq:micro_geometric}
  P(\x | \mu) = \multiset{N}{\mu}^{-1}\delta_{\mu,\sum_ix_i},
\end{equation}
where $\multiset{N}{\mu}={N+\mu-1\choose \mu}$ counts the number of ways
to distribute a total of $\mu$ values into $N$ distinguishable
parts. The marginal distribution of the individual covariates $x_i$ can
be obtained as
\begin{align}
  P(x_i | N, \mu) &= \frac{P(\x|\mu)}{P(\x\setminus x_i|\mu-x_i)},\\
   &= \frac{(N+\mu-x_i-1)!\mu!}{(N+\mu-1)!(\mu-x_i)!}H(\mu-x_i). \label{eq:geom_marginal}
\end{align}
Like with the continuous model, for sufficiently large $N$ and with the
mean $\mu/N$ fixed, the marginal distribution of individual values $x_i$
will follow asymptotically a geometric distribution with $p=N/(\mu + N)$
(see Fig.~\ref{fig:geom}).

Since the value of the parameter $\mu$ is also non-negative, we can
sample it from the same distribution as a prior. Putting this in the SBM
yields
\begin{equation}
  P(\x | \A, \bm{\mu}^1, \bb^1) = \prod_{r\le s}P(\x_{rs}^1|\mu_{rs}^1)
\end{equation}
and the final marginal distribution
\begin{widetext}
\begin{equation}\label{eq:marg_w_nested_geometric}
  P(\x | \A,\{\bb^l\})
  = \sum_{\{\mu^l_{rs}\}}P(\x | \A, \bm{\mu}^1, \bb^1)\prod_{l=1}^L\prod_{r\le s}\left[P(\bm\mu^l_{rs}|\mu^{l+1}_{b_r^{(l+1)},b_s^{(l+1)}})\right]^{1-\delta_{m_{rs}^l,0}}\\
  = \prod_{l=1}^L\prod_{r\le s}\left[\multiset{m_{rs}^l}{\bar\mu_{rs}^l}^{-1}\right]^{1-\delta_{\bar\mu_{rs}^l,0}},
\end{equation}
\end{widetext}
with $\bar\mu_{rs}^l$ given by Eq.~\ref{eq:bar_mu}. Like with the
continuous models, the use of Eq.~\ref{eq:marg_w_nested_geometric}
requires only a simple modification of the algorithm of
Ref.~\cite{peixoto_nonparametric_2017}, that does not alter its
algorithmic complexity.

\subsection{Binomial discrete weights}

Often, discrete covariates are bounded in a finite range
$x\in\{0,\dots,M\}$ (a common example are ratings in recommendation
systems~\cite{godoy-lorite_accurate_2016}). In this case, the
appropriate distribution is the binomial,
\begin{equation}
  P(x|p,M) = {M\choose x}p^x(1-p)^{M-x},
\end{equation}
where the value $x$ is commonly interpreted as the sum of $M$
independent Bernoulli outcomes with a probability $p\in[0,1]$ of
success. Incorporating it in the SBM yields
\begin{multline}
  P(\x_{rs} | \A, \bb, \bm{p},N)\\
\begin{aligned}
  \quad &= \prod_{ij}P(x_{ij}|p_{rs},N)^{\frac{A_{ij}\delta_{b_i,r}\delta_{b_i,r}}{(1+\delta_{rs})}} \\
  &= \left[\prod_{ij}{M\choose x_{ij}}^{\frac{A_{ij}\delta_{b_i,r}\delta_{b_i,r}}{(1+\delta_{rs})}}\right] p_{rs}^{\mu_{rs}}(1-p_{rs})^{Mm_{rs}-\mu_{rs}}.
\end{aligned}
\end{multline}
The conjugate prior is the beta distribution of Eq.~\ref{eq:beta} again,
yielding the marginal after integration over all $p_{rs}$,
\begin{multline}
  P(\x|\A, \bb, \alpha, \beta) =\\
  \left[\prod_{i<j}{M\choose x_{ij}}\right]\prod_{r\le s}\frac{B(\mu_{rs}+\alpha, Mm_{rs}-\mu_{rs}+\beta)}{B(\alpha,\beta)}.
\end{multline}
Once more, we can make the uninformative choice $\alpha=\beta=1$, which yields
\begin{align}
  P(\x|\A, \bb) &=
  \left[\prod_{i<j}{M\choose x_{ij}}\right]\prod_{r\le s}\frac{\mu_{rs}!(Mm_{rs}-\mu_{rs})!}{(Mm_{rs}+1)!}.
\end{align}
But as for the other cases, the best path for a hierarchical model is
through a microcanonical model, as described in the following.

\subsubsection{Microcanonical distribution}

\begin{figure}
  \includegraphics[width=\columnwidth]{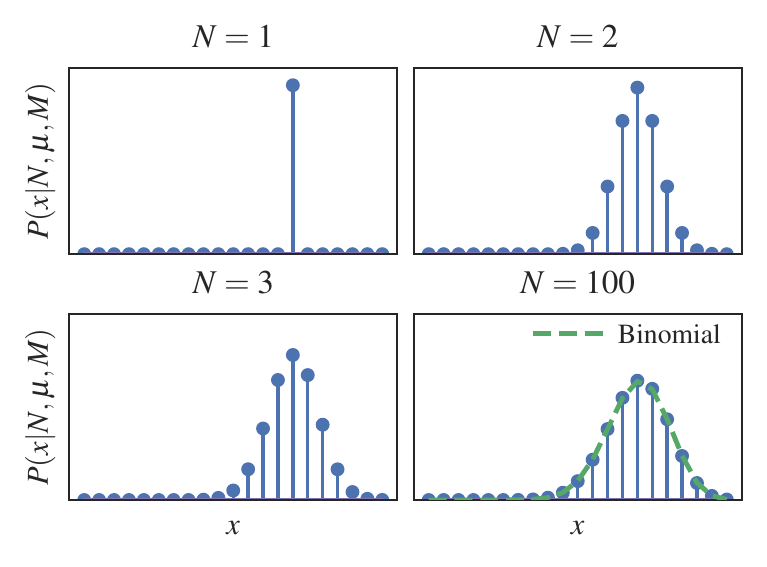}

  \caption{The marginal distribution of each individual covariate $x$ in
  the discrete microcanonical model, given by
  Eq.~\ref{eq:binom_marginal}, approaches asymptotically the binomial
  distribution as the number of values $N$ increases, and the mean
  $\bar{x}=\mu/N$ is kept fixed.\label{fig:binomial}}
\end{figure}

A microcanonical version of the Binomial distribution --- i.e. the
uniform distribution of $N$ non-negative discrete values $\x$, where
each value is bounded in the range $x_i\in\{0,\dots,M\}$, conditioned in
the total sum $\mu$ --- can be obtained by randomly sampling exactly
$\mu$ positive outcomes from a total of $NM$ trials. The joint
probability for $\x=\{x_1,\dots,x_N\}$ is therefore
\begin{equation}
  P(\x | \mu, M) = \left[\prod_i{M\choose x_i}\right]{MN\choose \mu}^{-1}\delta_{\mu,\sum_ix_i},
\end{equation}
where ${MN\choose \mu}$ counts the possible distributions of $\mu$
positive outcomes of $MN$ distinguishable trials, and the remaining terms
discount all outcomes that lead to the same value of $\x$.
The marginal distribution of the individual covariates $x_i$ can be
obtained as
\begin{multline}
  P(x_i |N,\mu,M)\\
  \begin{aligned}
    &= \frac{P(\x|\mu,M)}{P(\x\setminus x_i|\mu-x_i,M)},\\
    &= {M\choose x_i} \frac{[M(N-1)]![M(N-1)-\mu+x_i]!\mu!}{(MN)!(\mu-x_i)!(MN-\mu)!}
    \label{eq:binom_marginal}.
  \end{aligned}
\end{multline}
Like with the previous models, for sufficiently large $N$ and with
$\mu/N$ fixed, the marginal distribution of individual values $x_i$ will
follow asymptotically a binomial distribution with $p=\mu/(NM)$ (see
Fig.~\ref{fig:binomial}).

The parameter $\mu$ is a non-negative integer that can be chosen
arbitrarily, as long as the inequality $M \ge \mu/N$ is
satisfied. Therefore, we may sample $\mu$ from the distribution of
Eq.~\ref{eq:micro_geometric} in an unconstrained manner, and then sample
the parameter $M$ from a constrained distribution $P(M|\mu, N)$.
Incorporating this into the SBM yields,
\begin{equation}
  P(\x | \A, \bm{\mu}^1, \bb^1) = \prod_{r\le s}P(\x_{rs}^1|\mu_{rs}^1, M),
\end{equation}
and the overall marginal distribution
\begin{widetext}
\begin{align}
  P(\x,M | \A,\{\bb^l\})
  &= \sum_{\{\mu^l_{rs}\}}P(\x | \A, \bm{\mu}^1, \bb^1)P(M|\bm{\mu}^1,\A,\bb^1)\prod_{l=1}^L\prod_{r\le s}\left[P(\bm\mu^l_{rs}|\mu^{l+1}_{b_r^{(l+1)},b_s^{(l+1)}})\right]^{1-\delta_{m_{rs}^l,0}}\\
  &= P(M|\bar{\bm{\mu}}^1,\A,\bb^1)\left[\prod_{i\le j}{M\choose x_{ij}}\right]\left[\prod_{r\le s}{Mm_{rs}^1\choose \bar\mu_{rs}^1}^{-1}\right]\prod_{l=2}^L\prod_{r\le s}\left[\multiset{m_{rs}^l}{\bar\mu_{rs}^l}^{-1}\right]^{1-\delta_{\bar\mu_{rs}^l,0}}\label{eq:marg_w_nested_binomial},
\end{align}
\end{widetext}
where $P(M|\bar{\bm{\mu}}^1,\A,\bb^1)$ is a prior distribution for $M$
that respects the constraint $M\ge \bar\mu_{rs}^1/m_{rs}^1$. Thus, given
any arbitrary value $M^*$, we can choose
\begin{multline}
  P(M|\bar{\bm{\mu}}^1,\A,\bb^1)\\
  = \begin{cases}
    1 & \text{ if } M = \max\left(M^*, \ceil{\max_{rs}\bar\mu^1_{rs}/m^1_{rs}}\right),\\
    0 & \text{ otherwise,}
  \end{cases}
\end{multline}
such that if $M^*$ is compatible with the observed covariates,
i.e. $M^*\ge x_{ij}$, we have $P(M^*|\bar{\bm{\mu}}^1,\A,\bb^1)=1$ for
any possible value of $\bar{\bm{\mu}}^1$ and $\bb^1$ encountered in the
posterior, as long as $M=M^*$, thereby effectively removing it from
Eq.~\ref{eq:marg_w_nested_binomial}.  A completely nonparametric
approach would require us to include a prior $P(M^*)$, but since it is a
single global number, we can safely omit it, as it cannot influence the
posterior distribution of partitions. In most practical scenarios, the
bound $M^*$ is known \emph{a priori}; otherwise it can be chosen as
$M^*=\max_{ij}x_{ij}$.

\subsection{Poisson discrete weights}

A natural extension of the binomial weights is the situation where
$M\to\infty$ with the mean $\lambda = pM$ kept fixed, which yields
the Poisson distribution
\begin{equation}
  P(x|\lambda) = \frac{\lambda^x \ee^{-\lambda}}{x!}.
\end{equation}
Using this in the SBM gives us
\begin{align}
  P(\x | \A, \bb, \bm{\lambda})
  &= \prod_{i<j}P(x_{ij}|\lambda_{b_i,b_j})^{A_{ij}} \\
  &= \left[\prod_{i<j}x_{ij}!^{A_{ij}}\right]^{-1}\prod_{r\le s} \lambda_{rs}^{\mu_{rs}}\ee^{-m_{rs}\lambda_{rs}}.
\end{align}
Once more, the conjugate prior is the gamma distribution of
Eq.~\ref{eq:gamma}, which after integrating over $\lambda_{rs}$ yields
the marginal distribution
\begin{multline}
  P(\x|\A, \bb, \alpha,\beta) =\\
  \left[\prod_{i<j}x_{ij}!^{A_{ij}}\right]^{-1}\prod_{r\le s}
  \frac{\beta^\alpha\Gamma(\mu_{rs}+\alpha)}{\Gamma(\alpha)(m_{rs}+\beta)^{\mu_{rs}+\alpha}}.
\end{multline}
The uninformative maximum entropy choice is $\alpha=1$, yielding
\begin{equation}
  P(\x|\A, \bb, \beta) =
  \left[\prod_{i<j}x_{ij}!^{A_{ij}}\right]^{-1}\prod_{r\le s}
  \frac{\beta \mu_{rs}!}{(m_{rs}+\beta)^{\mu_{rs}+1}}.
\end{equation}
But once more, we can obtain a deeper hierarchical model by formulating
an asymptotically equivalent microcanonical model.

\subsubsection{Microcanonical distribution}

\begin{figure}
  \includegraphics[width=\columnwidth]{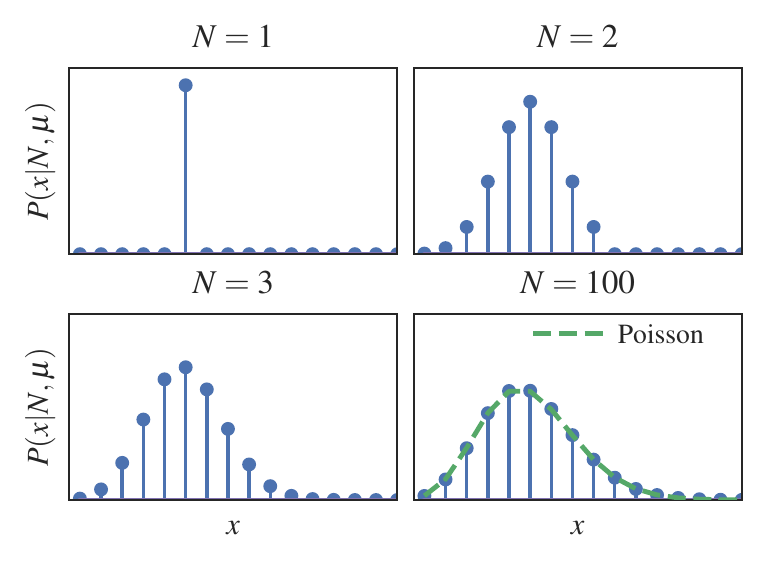}

  \caption{The marginal distribution of each individual covariate $x$ in
  the discrete microcanonical model, given by Eq.~\ref{eq:marg_poisson},
  approaches asymptotically the Poisson distribution as the number of
  values $N$ increases, and the mean $\bar{x}=\mu/N$ is kept
  fixed.\label{fig:poisson}}
\end{figure}

The joint distribution of $N$ Poisson variables $\x=\{x_1,\dots,x_N\}$
can be decomposed into a Poisson distribution for the total sum $\mu$
with mean $N\lambda$ and a uniform multinomial distribution for $\x$
conditioned on the total sum, i.e.
\begin{equation}
  P(\x | \lambda) = P(\x | \mu) P(\mu|N\lambda).
\end{equation}
The microcanonical version, therefore, is given simply by replacing
$P(\mu|N\lambda)\to\delta_{\mu,\sum_ix_i}$, yielding
\begin{equation}
  P(\x | \mu) = \frac{\mu!}{\prod_ix_i!}\frac{1}{N^{\mu}}\delta_{\mu,\sum_ix_i}.
\end{equation}
The marginal distribution of the individual covariates $x_i$ can be
obtained again as
\begin{align}
  P(x_i |N,\mu) &= \frac{P(\x|\mu)}{P(\x\setminus x_i|\mu-x_i)},\\
  &= \frac{\mu!(N-1)^{\mu-x_i}}{(\mu-x)!N^{\mu}x_i!},
  \label{eq:marg_poisson}
\end{align}
The global constraint on the total sum has a vanishing effect for
sufficiently large $N$, as long as the mean $\mu/N$ is kept fixed, as
the marginal distribution of individual values $x_i$ will follow
asymptotically a Poisson distribution with $\lambda=\mu/N$ (see
Fig.~\ref{fig:poisson}).

The parameter $\mu$ is a non-negative integer that can be chosen
arbitrarily. Therefore, we may sample $\mu$ from the distribution of
Eq.~\ref{eq:micro_geometric}. Incorporating this into the SBM yields,
\begin{equation}
  P(\x | \A, \bm{\mu}^1, \bb^1) = \prod_{r\le s}P(\x_{rs}^1|\mu_{rs}^1),
\end{equation}
and the overall marginal distribution
\begin{widetext}
\begin{align}
  P(\x | \A,\{\bb^l\})
  &= \sum_{\{\mu^l_{rs}\}}P(\x | \A, \bm{\mu}^1, \bb^1)\prod_{l=1}^L\prod_{r\le s}\left[P(\bm\mu^l_{rs}|\mu^{l+1}_{b_r^{(l+1)},b_s^{(l+1)}})\right]^{1-\delta_{m_{rs}^l,0}}\\
  &= \left[\prod_{i\le j}x_{ij}!^{A_{ij}}\right]^{-1}\left[\prod_{r\le s}\left(\frac{\bar\mu_{rs}^1!}{(m_{rs}^1)^{\bar\mu_{rs}^1}}\right)^{1-\delta_{m_{rs}^1,0}}\right]\prod_{l=2}^L\prod_{r\le s}\left[\multiset{m_{rs}^l}{\bar\mu_{rs}^l}^{-1}\right]^{1-\delta_{\bar\mu_{rs}^l,0}}.\label{eq:marg_w_nested_poisson}
\end{align}
\end{widetext}

\subsection{Transformed weights}\label{sec:transformations}

The models above can be easily modified to accommodate a much wider
class of covariates, without any substantial change to the likelihoods,
via variable transformations of the type $y_{ij}=f(x_{ij})$, according
to some function $f(x)$. For the continuous models in particular, such
variable transformations yield the scaled marginal likelihoods
\begin{equation}
  P(\x | \A,\{\bb^l\}) =  P(\bm{y}(\x) | \A,\{\bb^l\}) \prod_{i<j}\left[\frac{\dd f}{\dd x}(x_{ij})\right]^{A_{ij}}.
\end{equation}
The product of derivatives in the equation above is a multiplicative
constant that does not depend on the hierarchical partition $\{\bb^l\}$,
and hence does not affect the posterior distribution (although it is
relevant for model selection; see Sec.~\ref{sec:model-selection}
below). We are thus free to choose any weight transformation $f(x)$, and
use the previously defined distributions and associated algorithms on
the transformed weights, without any other alteration. This gives us a
wider class of covariate models that may be better suitable for specific
datasets, and can be developed in an \emph{ad hoc} manner. In the
following, we cover some typical examples, non-exhaustively.

\subsubsection{Broadly distributed weights}

If the observed weights are positive and broadly distributed, a possibly
better model is the Pareto distribution,
\begin{equation}
  P(x|\alpha, x_m) = \begin{cases}
    \displaystyle\frac{\alpha x_m^\alpha}{x^{\alpha+1}},& \text{ if } x > x_m,\\
    0, & \text{ otherwise.}
    \end{cases}
\end{equation}
Instead of computing the integrated likelihood from scratch, we use the
fact that the variable transformation $y = \ln(x/x_m)$ yields
\begin{equation}
  P(y|\alpha) = \alpha \ee^{-\alpha y},
\end{equation}
which is the exponential distribution we used before. So when dealing
with broad weights, we can just make this transformation on the weights
and use the exponential model.

Alternatively, we may use the normal model for $y=\ln x$, which assumes
that $x$ is distributed according to a log-normal. In our experience, we
found that this choice also typically yields better results when the
positive weights are peaked around a typical value, in a manner that is
difficult to represent with a mixture of exponential distributions.

\subsubsection{Bounded weights}

If the weights are bounded in an interval $x \in [a, b]$, we can adapt
it to an unbounded distribution by first uniformly mapping the weights
to the unit interval $x'\in[0,1]$, via
\begin{equation}\label{eq:unit}
  x' = \frac{x-a}{b-a},
\end{equation}
and then using a logit transformation
\begin{equation}
  y = \ln\left(\frac{x'}{1-x'}\right),
\end{equation}
or, equivalently, first mapping to the symmetric interval $x'\in[-1,1]$, via
\begin{equation}
  x' = 2\frac{x-a}{b-a}-1,
\end{equation}
and using the inverse hyperbolic tangent
\begin{equation}
  y = 2\,\text{arctanh}(x') = \ln\left(\frac{1+x'}{1-x'}\right),
\end{equation}
both of which yield the same signed unbounded weight $y \in [-\infty,
\infty]$, which can be fit using the normal distribution.
Alternatively, the negative logarithm can be used with Eq.~\ref{eq:unit}
\begin{equation}
  z = -\ln x',
\end{equation}
which yields a positive unbounded weight $z \in [0, \infty]$ that can be
used with the exponential distribution.  Which approach is most suitable
depends on the actual shape of the data, and can be determined \emph{a
posteriori} via model selection, as described in
Sec.~\ref{sec:model-selection}.

\subsubsection{Decomposing covariates}

We can also obtain more elaborate models by decomposing a single
covariate into multiple ones. Consider, for example, the case of signed
discrete weights $x\in [\dots, -2, -1, 0, 1, 2,\dots]$, which was not
considered directly by any of the models so far. This can be done in a
straightforward manner by decomposing the numbers into a sign and
magnitude, i.e.
\begin{equation}
  x_{ij} = (2s_{ij}-1)y_{ij}
\end{equation}
where
\begin{align}
  s_{ij} &= (\operatorname{sgn}(x_{ij}) + 1)/2,\\
  y_{ij} &= \operatorname{abs}(x_{ij})
\end{align}
is a reversible transformation that extracts the sign and absolute
values of $x_{ij}$. We may then use a Binomial distribution with $M=1$
(i.e. Bernoulli) for $s_{ij}\in\{0,1\}$, and any non-negative
distribution for $y_{ij}\in\{0,1,2,\dots\}$, and obtain the posterior
using the joint marginal likelihood
\begin{align}
  P(\x|\A,\{\bb^l\}) &= P(\bm{y},\bm{s}|\A,\{\bb^l\})\\
  &= P(\bm{y}|\A,\{\bb^l\})P(\bm{s}|\A,\{\bb^l\}).
\end{align}

\subsection{Model selection}\label{sec:model-selection}

Given any two models $\mathcal{M}_1$ and $\mathcal{M}_2$ for the same
weighted network with edge covariates $\x$, for which we obtain the
partitions $\{\bb^l\}_1$ and $\{\bb^l\}_2$ from their respective
posterior distributions, we can perform model selection as described in
Ref.~\cite{peixoto_nonparametric_2017}, by computing the posterior odds
ratio
\begin{align}
  \Lambda &= \frac{P(\{\bb^l\}_1,\mathcal{M}_1|\A,\x)}{P(\{\bb^l\}_2,\mathcal{M}_2|\A,\x)}\\
          &= \frac{P(\A|\{\bb^l\}_1,\mathcal{M}_1)P(\x|\A,\{\bb^l\}_1,\mathcal{M}_1)P(\{\bb^l\}_1)P(\mathcal{M}_1)}{P(\A|\{\bb^l\}_2,\mathcal{M}_2)P(\x|\A,\{\bb^l\}_2,\mathcal{M}_2)P(\{\bb^l\}_2)P(\mathcal{M}_2)},
\end{align}
where $P(\mathcal{M})$ is the prior preference for either model
[typically, we are agnostic with
$P(\mathcal{M}_1)=P(\mathcal{M}_2)$]. For values of $\Lambda > 1$, the
choice $(\{\bb^l\}_1,\mathcal{M}_1)$ is preferred over
$(\{\bb^l\}_2,\mathcal{M}_2)$ according to the data, and the magnitude
of $\Lambda$ yields the degree of statistical significance.

Using this criterion we can select between unweighted variations of the
SBM (e.g. degree-corrected or not)~\cite{peixoto_nonparametric_2017},
but also between different models of the weights. This is particularly
useful when using weight transformations as described in
Sec.~\ref{sec:transformations}. For example, when considering two
different transformations $y_{ij}=f(x_{ij})$ and $z_{ij}=g(x_{ij})$,
using different models $\mathcal{M}_y$ and $\mathcal{M}_z$ for the
transformed covariates, the posterior odds ratio [with
agnostic priors $P(\mathcal{M}_y)=P(\mathcal{M}_z)$] becomes
\begin{equation}
  \Lambda = \frac{P(\A,\bm{y}(\x)|\A,\{\bb^l\}_1,\mathcal{M}_y)P(\{\bb^l\}_1)\prod_{i<j}f'(x_{ij})^{A_{ij}}}{P(\A,\bm{z}(\x)|\A,\{\bb^l\}_2,\mathcal{M}_z)P(\{\bb^l\}_2)\prod_{i<j}g'(x_{ij})^{A_{ij}}}.
\end{equation}
The approach is entirely analogous for transformations on discrete
weights, where one simple omits the derivative terms.

\begin{table}
  \begin{tabular}{llll}
    Transformation & Derivatives & Weight model & $\ln P(\A,\x,\{\bb^l\})$\\ \hline\hline
    $y_{ij}=x_{ij}$ & $1$ & Exponential &  $-56,512$\\
    $y_{ij}=\ln x_{ij}$ & $\prod_{i<j}1/x_{ij}^{A_{ij}}$ & Normal & $-52,054$\\
  \end{tabular}

  \caption{Joint log-likelihood $\ln P(\A,\x,\{\bb^l\})$ for the human
  brain data in Sec.~\ref{sec:brain} using the electrical connectivity
  as edge covariate, for two model variations according to weight
  transformations. \label{tb:model-selection}}
\end{table}

\begin{figure}
  \includegraphics[width=\columnwidth]{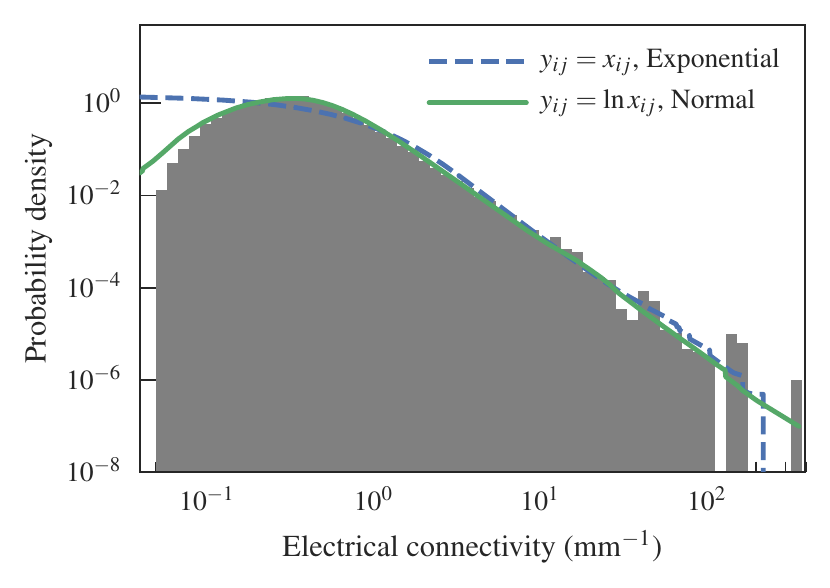}

  \caption{Overall distribution of the electrical connectivity of the
  human brain data. The solid lines shows the inferred distribution
  according to the weighted SBM using two models for the edge
  covariates, as shown in the legend.\label{fig:model-selection}}
\end{figure}

We illustrate the use of this criterion on the human brain data analyzed
in Sec.~\ref{sec:brain}. We consider here only the electric connectivity
covariate, which is non-negative and unbounded in the range
$[0,\infty]$. We consider two models for the weights: The first is the
exponential model of Sec.~\ref{sec:expon} applied directly to the
original covariates, and the second is the normal model of
Sec.~\ref{sec:normal}, applied to the transformed weights
$y_{ij}=\ln x_{ij}$, which results in a log-normal model for
$x_{ij}$. As the results of Table~\ref{tb:model-selection} show, we
obtain for this dataset a posterior odds ratio of $\ln\Lambda\approx
4,458$ favoring the log-normal model, despite the fact that it contains
more internal parameters. As we see in Fig.~\ref{fig:model-selection},
indeed the log-normal model is better suited to capture the peaked
nature of the overall distribution. It should be noted that while it is
a trivial feat to obtain better fits with more complicated models, the
Bayesian criterion above takes into account the complexity of the model,
and will point towards a more complicated one only if the statistical
evidence in the data supports it.

\section{Conclusion}\label{sec:conclusion}

The weighted extensions of the SBM presented in this work allow for a
principled inference of large-scale modular structure of weighted
networks, in a manner that is fully nonparametric, and algorithmically
efficient. As they include a hierarchical description of the network ---
taking into account both the node adjacency as well as the edge weights
--- our SBM implementations enable the detection of modular structures
at multiple scales, without being biased towards any specific kind of
mixing pattern (such as assortativity) in any of them.

The nonparametric nature of our approach means that it can be used to
detect the most appropriate model dimension, including the number of
groups as well as size and shape of the hierarchical division, directly
from data, in a parsimonious way, without requiring any prior
input. This comes with the guarantee that the inferred hierarchy is
statistically significant, and hence is not the result of statistical
fluctuations of a simpler model (such as a completely random graph).

The edge weights are included in the model description as additional
covariates, and thus require specific models that reflect their
nature. The explicit variations presented in this work cover a broad
range of possible types of covariates, that can be either continuous or
discrete, signed or unsigned, bounded or unbounded. Furthermore, all
these particular variations can be arbitrarily extended to accommodate a
much wider class of weight models via variable transformations, which
incur no modification to the algorithms. Such transformations can be
performed in an \emph{ad hoc} manner, reflecting the specificity of the
data at hand, and the best choice can be evaluated \emph{a posteriori}
using Bayesian model selection, simultaneously taking into account the
quality of fit, the model complexity and the statistical evidence
available from the data.

Although we do not describe this in detail here, it is easy to see that
the exact same approach we present can be used for other variations of
the SBM, such as with overlapping
groups~\cite{ball_efficient_2011,peixoto_model_2015}, edge
layers~\cite{peixoto_inferring_2015,stanley_clustering_2016,heimlicher_community_2012} and dynamic
networks~\cite{xu_dynamic_2013,peixoto_inferring_2015,peixoto_modelling_2017}.

Despite its advantages, our approach inherits the limitations of the
underlying SBM ansatz. In particular, it assumes that the weights are
distributed on the edges in a manner that is (asymptotically, in the
microcanonical case) conditionally independent. Hence, in the same
manner that the unweighted SBM does not include the often realistic
propensity of the network to form triangles and other local structures,
the weighted extensions preclude the existence of certain kinds of
weight correlations that are known to exist in key
cases~\cite{macmahon_community_2015}. The development of tractable and
versatile models that incorporate such higher-order aspects remains an
open challenge.

\appendix
\vspace{1em}
\section{Directed networks}\label{app:directed}
\vspace{-1em}

Although we focused on undirected networks in the main text, our methods
can be easily adapted to directed networks. The models for directed
adjacency matrices $P(\A|\{\bb^l\})$ are described in detail in
Ref~\cite{peixoto_nonparametric_2017}. For the edge covariates, the
modifications are straightforward yielding expressions for
$P(\x|\A,\{\bb^l\})$ that are identical, but with products going over
directed pairs of groups and nodes, i.e. $\prod_{r\le s}\to\prod_{rs}$
and $\prod_{i\le j}\to\prod_{ij}$. Our reference implementation supports
these variations~\cite{peixoto_graph-tool_2014}.

\bibliography{bib}

\end{document}